\documentclass[letterpaper, 10 pt, journal]{IEEEtran}
\IEEEoverridecommandlockouts

\usepackage[T1]{fontenc}
\usepackage{times}
\usepackage{microtype}
\usepackage{amsmath,amssymb}
\usepackage{graphicx,xcolor}
\usepackage{capt-of}
\usepackage{booktabs,tabularx,multirow}
\usepackage{xspace}
\usepackage[hidelinks]{hyperref}
\usepackage[capitalize]{cleveref}
\crefname{figure}{Fig.}{Figs.}
\Crefname{figure}{Fig.}{Figs.}
\ifdefined\pdfminorversion\pdfminorversion=7\fi

\makeatletter
\newcommand{\authornotes}{}
\newcommand{\deferauthornote}[1]{\g@addto@macro\authornotes{\footnotetext[0]{#1}}}
\newcommand{\printauthornotes}{
  \begingroup
  \let\@makefnmark\relax
  \authornotes
  \endgroup}
\makeatother

\newcommand{\dataset}{\textup{EoL3D}\xspace}
\title{\LARGE \bfseries \dataset: A Dataset and Benchmark for End-of-Life \\ Desktop Disassembly}
\author{
  Chang~Liu$^{1}$\quad Haiyang~Zhao$^{2}$\quad Jiurun~Song$^{1}$\quad Badrinath~Balasubramaniam$^{2}$\\[0.05em]
  Jiachen~Li$^{3}$\quad Zhiwen~Fan$^{1}$\quad Sara~Behdad$^{4}$\quad Yu~She$^{5}$\quad
  Beiwen~Li$^{2}$\quad Xiao~Liang$^{1,*}$\quad Minghui~Zheng$^{1,*}$\\[0.8em]
  {\normalsize $^{1}$Texas A\&M University\quad $^{2}$University of Georgia\quad $^{3}$Georgia Institute of Technology}\\
  {\normalsize $^{4}$University of Florida\quad $^{5}$Purdue University}\thanks
  {This work was partially supported by the U.S. National Science Foundation under Grant Nos. 2026533/2422826, 2132923/2422640, 2026276, 2132773/2506209. Portions of this research were conducted with the advanced computing resources provided by Texas A\&M High Performance Research Computing. Correspondence to Xiao Liang and Minghui Zheng (\href{mailto:xliang@tamu.edu}{\nolinkurl{xliang@tamu.edu}}, \href{mailto:mhzheng@tamu.edu}{\nolinkurl{mhzheng@tamu.edu}}).}}

\IEEEaftertitletext{\vspace{-0.9em}
  \noindent\begin{minipage}{\textwidth}
    \centering
    \vspace{-1.2em}
    {\textbf{Project page:} \href{https://eol3d.github.io/}{\bf \textcolor{blue}{\nolinkurl{https://eol3d.github.io/}}}\par}
    \vspace{1em}
    \includegraphics[width=0.98\linewidth]{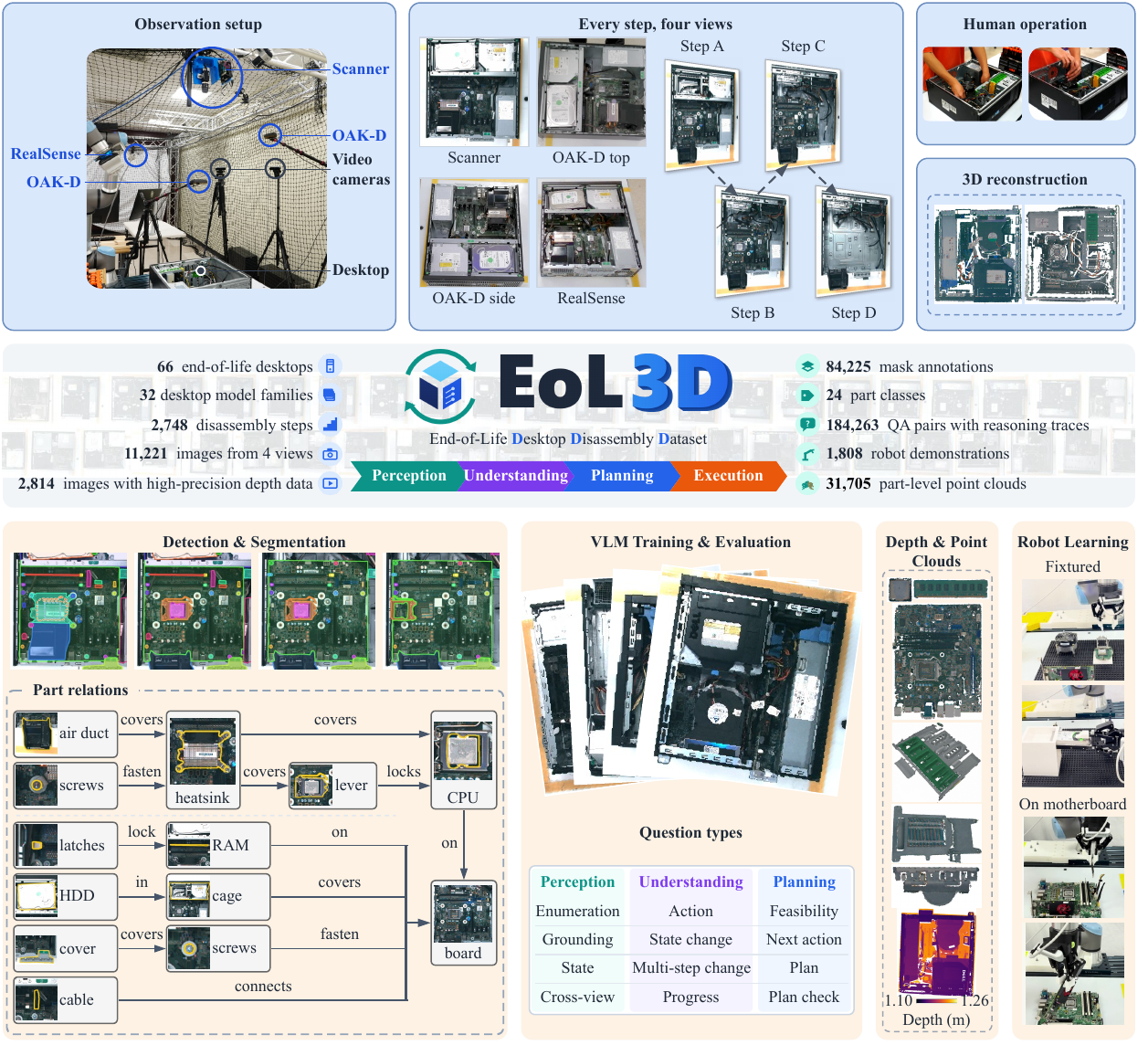}\par
    \captionof{figure}{\textbf{Overview of \dataset.} EoL desktops are disassembled under a fixed observation setup. After every step, the remaining assembly is imaged from four views and reconstructed in 3D, and two side-view cameras record the human operation. EoL3D provides scanner-view annotations that link persistent instance identities with visible and full-extent masks, component states, and the part relations that define removal prerequisites. These records support component detection and segmentation, vision-language questions on perception, understanding, and planning, and depth and part-level point-cloud data. Separately collected robot demonstrations cover component removal in the fixture and on-motherboard settings.}
    \label{fig:overview}
  \end{minipage}\par
  \vspace{0.35em}
}

\begin{document}
\onecolumn
\let\thanks\deferauthornote
\maketitle
\thispagestyle{empty}
\pagestyle{empty}
\raggedbottom
\twocolumn
\printauthornotes

\begin{abstract}

End-of-life (EoL) desktop disassembly requires understanding the product's current state, identifying and locating components, and determining subsequent operations for both computer-assisted human disassembly and robotic planning and execution. Collected EoL desktops often vary across brands and models, and their internal configurations and conditions are not fully known beforehand, so decisions must be updated at each step as disassembly reveals hidden components and changes access to others. Studying these capabilities requires data that follow individual components and their changing states throughout disassembly processes and connect new observations with completed operations. We introduce the EoL Desktop Disassembly Dataset (EoL3D), which documents 2,748 disassembly steps with images from four views across 66 EoL desktops from different brands, with various internal configurations, and under different usage conditions. EoL3D maintains component identities across disassembly steps and links visible and full-extent masks with component states and removal prerequisites. The dataset also includes stepwise 3D reconstructions of the remaining assembly, part-level point clouds, two-view videos segmented by operation, over 180k QA pairs, and 1,808 separately collected robot demonstrations of component removal. To support different research tasks, we establish benchmarks for component detection, instance segmentation, and monocular depth estimation for the computer vision domain. QA pairs are used to benchmark state-of-the-art closed frontier models and open-source vision-language models (VLMs) on their perception, understanding, and planning capabilities. We also benchmark the vision-language-action (VLA) policies for complex robotic disassembly manipulation.
\end{abstract}

\section{Introduction}
\label{sec:introduction}

The growing volume of end-of-life (EoL) products has drawn increasing attention to remanufacturing and recycling~\cite{liu2027transforming}. Disassembly is an initial and central step in both processes, where the value and subsequent processing of collected EoL products must be determined. However, the condition uncertainty and configuration variety are the natures that are associated with the collected EoL products, requiring adaptive real-time decision-making and replanning during the disassembly processes~\cite{tian2026redefining}. Currently, disassembly relies heavily on human expertise to handle these challenges. However, the cost of human labor makes the disassembly practice unprofitable and misaligned with the growing volume of EoL products, encouraging automated robotic disassembly to improve efficiency~\cite{liu2026raise}.

Driven by large-scale datasets, recent advances in robotic learning have substantially improved the end-to-end manipulation performance~\cite{walke2023bridgedata, khazatsky2024droid, o2024open}. However, most of the work focuses on daily-life manipulation tasks, such as pickup and place, rearranging objects, and packing~\cite{kim2024openvla, intelligence2025pi_}. While some of them include long-horizon in-house tasks~\cite{intelligence2026pi}, industrial disassembly involves additional complexity from the variety of operations and their sequential constraints~\cite{liu2026pacer}. In EoL desktops, component appearance, layout, and surrounding structures vary across brands and models. Accurate spatial information is needed to locate these components for precise manipulation. Each disassembly operation can expose hidden components and change the feasibility of different disassembly operations. The robot must understand the current state, identify the components, locate their positions, and determine the subsequent operations at each step of the disassembly process.

Existing e-waste datasets collect multi-view images during disassembly for screw detection and laptop component segmentation~\cite{kalitsios2022vision, zhang2026evaluating}. Their focus is on individual images rather than stepwise records of component removals and the remaining assembly. For robotic manipulation, REASSEMBLE provides multimodal robot demonstrations and performance benchmarks for assembly and disassembly on the standardized NIST task board~\cite{sliwowski2025reassemble}. IMPACT records human assembly and disassembly of two angle grinder models from synchronized egocentric and external views~\cite{wen2026impact}. It annotates actions and links them to component states and procedural prerequisites. However, these datasets provide limited coverage of EoL desktop disassembly across different configurations and conditions. We need a dataset that follows actual EoL desktop disassembly across different configurations and usage conditions. It needs to connect visual and spatial observations of components with their states and the feasible operations, supporting the transition from perception to decision-making, replanning, and guided execution at each disassembly step.

\begin{table*}[t]
\caption{Comparison with representative assembly and disassembly datasets.}
\label{tab:comparison}
\centering
\footnotesize
\setlength{\tabcolsep}{2.25pt}
\renewcommand{\arraystretch}{1.16}
\begin{tabularx}{\textwidth}{@{}
    >{\raggedright\arraybackslash}p{.142\textwidth}
    >{\raggedright\arraybackslash}p{.120\textwidth}
    >{\raggedright\arraybackslash}p{.303\textwidth}
    >{\raggedright\arraybackslash}X
    >{\raggedright\arraybackslash}p{.165\textwidth}
    @{}}
\toprule
\textbf{Dataset} & \textbf{Product domain} & \textbf{Data} & \textbf{Annotations} & \textbf{Scale} \\
\midrule
Assembly101~\cite{sener2022assembly101}
& Toy vehicles
& Multi-view video
& Actions, mistakes, 3D hand poses
& 4,321 videos \\
EgoProceL~\cite{bansal2022my}
& PCs (subset)
& Egocentric video
& Key steps
& 29 videos \\
IMPACT~\cite{wen2026impact}
& Angle grinders
& Multi-view RGB-D video
& Actions, states, prerequisites
& 112 trials \\
WDSD~\cite{kalitsios2022vision}
& Electronic waste
& RGB images
& 4,414~screw masks
& 945 images \\
Zhang et al.~\cite{zhang2026evaluating}
& MacBook laptops
& RGB images
& 1,717 component masks
& 560 raw images \\
RAISE~\cite{liu2026raise}
& Phones
& RGB images
& 6,939 layer masks
& 2,376 images \\
Al Assadi et al.~\cite{al2024towards}
& Robot vacuum
& Step images
& Operations, joining technologies
& 37 experiments \\
REASSEMBLE~\cite{sliwowski2025reassemble}
& NIST task board
& RGB, force/torque, audio, events, robot states
& Actions, skills, outcomes
& 4,551 action demos \\
\midrule
\textbf{\dataset\ (Ours)}
& EoL desktops
& Multi-view RGB-D, scans, video, 11,221~images, 31,705~part-level~point~clouds
& 84,225~masks, state representations, prerequisites, 184,263~QA pairs
& 66 desktops, 2,748~steps, and 1,808~robot demos \\
\bottomrule
\end{tabularx}
\vspace{-2pt}
\end{table*}

To bridge this gap, we introduce the EoL Desktop Disassembly Dataset (EoL3D) for learning and evaluating robotic disassembly. It records 2,748 disassembly steps across 66 distinct desktops from 32 model families with different internal configurations and usage conditions. After each operation, we capture the remaining assembly with three RGB-D cameras and an overhead high-precision 3D scanner with RGB imaging. We log the operation and tools used, while side-view videos record the full human disassembly process. Once disassembly is complete, we capture the removed components on a table using the same setup. We manually annotate the scanner-view images in reverse step order, using later observations to label components occluded in earlier steps. These annotations provide visible-region and full-extent masks, with component identities maintained across steps. Component relation graphs record occlusion and removal prerequisites. State labels track whether a component still constrains the removal of another. We reconstruct the remaining assembly point clouds from the scans and use the visible-region segmentation masks to extract part-level point clouds.

We benchmark component detection and visible-region instance segmentation using these annotations, and monocular depth estimation against depth maps derived from the scans. Using predefined question templates, we construct question-answer (QA) pairs from the component annotations and operation records. The questions assess perception, understanding, and planning through component recognition, state changes, and the feasibility and ordering of subsequent operations. We evaluate closed-source and open-source vision-language models (VLMs) in zero-shot settings and fine-tune open-source models on these pairs. To study robotic execution, we separately collect 1,808 component-removal demonstrations on fixtures and directly from motherboards. We use these demonstrations to train and evaluate vision-language-action (VLA) policies.

\section{Related Work}

\textbf{Assembly and Disassembly Datasets} Several existing disassembly datasets provide image annotations for screw detection~\cite{kalitsios2022vision, zhang2023automatic} and phone component detection~\cite{jahanian2019see}. Some datasets include instance segmentation annotations for laptop components from multiple views~\cite{zhang2026evaluating} and phone layers for sorting~\cite{liu2026raise}. Others focus on human manipulation during assembly and disassembly. Al Assadi et al.~\cite{al2024towards} record complete manual disassembly processes of EoL robot vacuum cleaners. Assembly101, IKEA ASM, and HA-ViD~\cite{sener2022assembly101,ben2021ikea,zheng2023ha} also provide video benchmarks for human action recognition. Considering the sequential constraints of assembly and disassembly, EgoProceL~\cite{bansal2022my} uses egocentric videos to study procedural steps and their order. IndustReal~\cite{schoonbeek2024industreal} uses assembly states to evaluate step completion. Recently, IMPACT~\cite{wen2026impact} has linked human actions to component states and operation prerequisites. With the rise of VLMs, EgoPlan-Bench~\cite{chen2026egoplan} evaluates their task planning capabilities based on the task goal, progress, and current observations. EoL3D includes component annotations, human operation videos, state labels, removal prerequisites, and QA pairs for desktop disassembly. Table~\ref{tab:comparison} compares EoL3D with representative datasets.

\textbf{Learning for Robotic Disassembly} Real-world robotic manipulation datasets such as DROID~\cite{khazatsky2024droid}, BridgeData V2~\cite{walke2023bridgedata}, and Open X-Embodiment~\cite{o2024open} provide demonstrations for learning robot policies across different tasks and environments. However, experiments on desktop disassembly show that VLA policies still struggle with precise contact operations after fine-tuning on disassembly demonstrations~\cite{liu2026vision}. For contact-rich assembly and disassembly, REASSEMBLE~\cite{sliwowski2025reassemble} provides robot demonstrations and policy benchmarks on the NIST task board. On the same task board, AlphaNIST~\cite{zhang2026alphanist} learns assembly policies through reinforcement learning in simulation using binary task rewards. To retain precision during contact operations, SELF-VLA~\cite{liu2026self} combines VLA policies with predefined disassembly skills in an agentic framework. The policies control the approach to components and determine when to invoke the skills. The skills execute precise contact operations, while learned policies handle failure recovery.

\section{Dataset}

\begin{figure*}[t]
\centering
\includegraphics[width=0.98\textwidth]{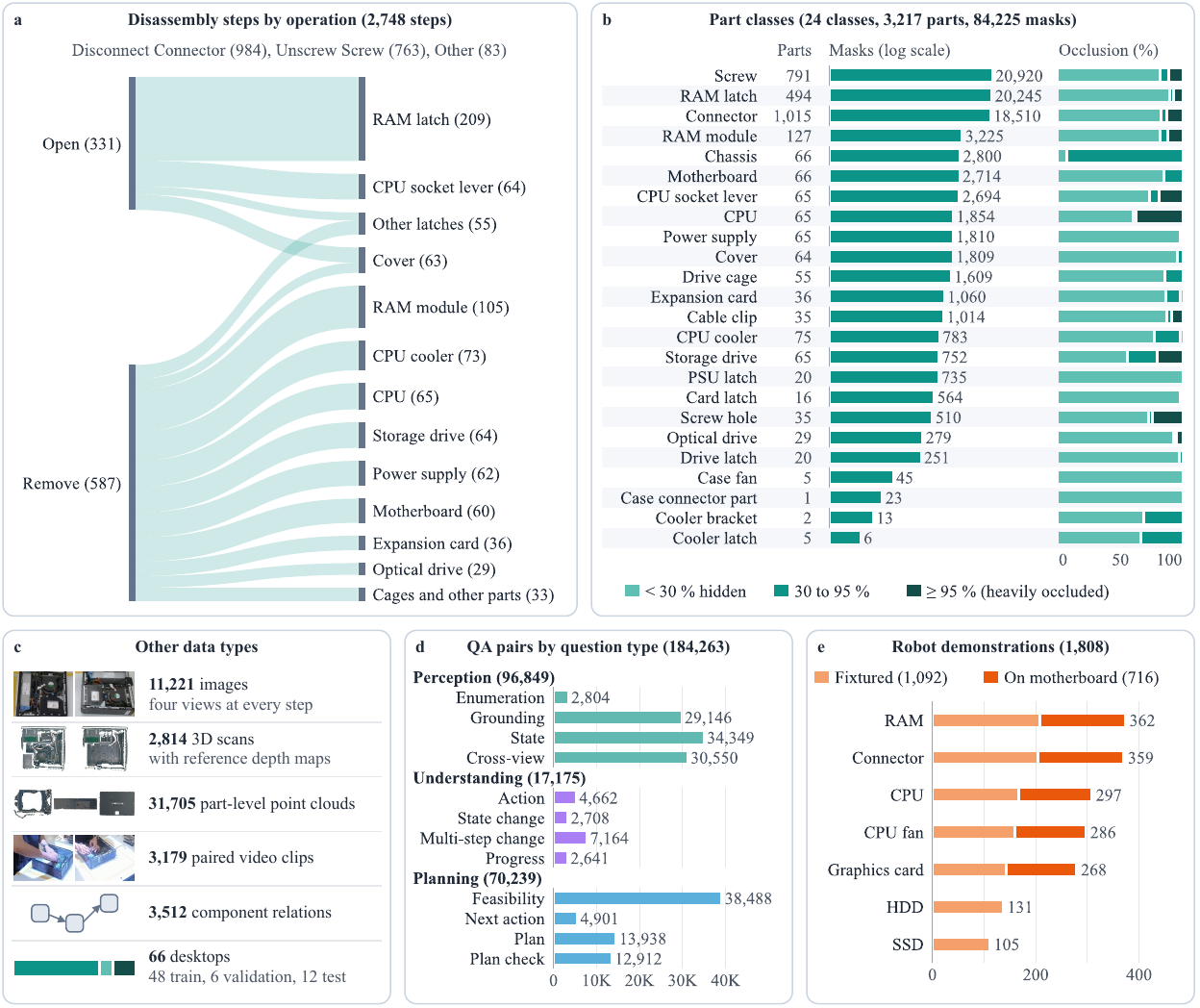}
\caption{\textbf{Statistics of \dataset.} (a) Disassembly steps by operation and target part. (b) Annotated parts and masks of each class, with their occlusion levels. (c) Other data types in the dataset. (d) QA pairs of each question type. (e) Robot demonstrations of each target component in the two collection settings.}
\label{fig:distributions}
\end{figure*}

\begin{figure*}[t]
    \centering
    \includegraphics[width=0.98\textwidth]{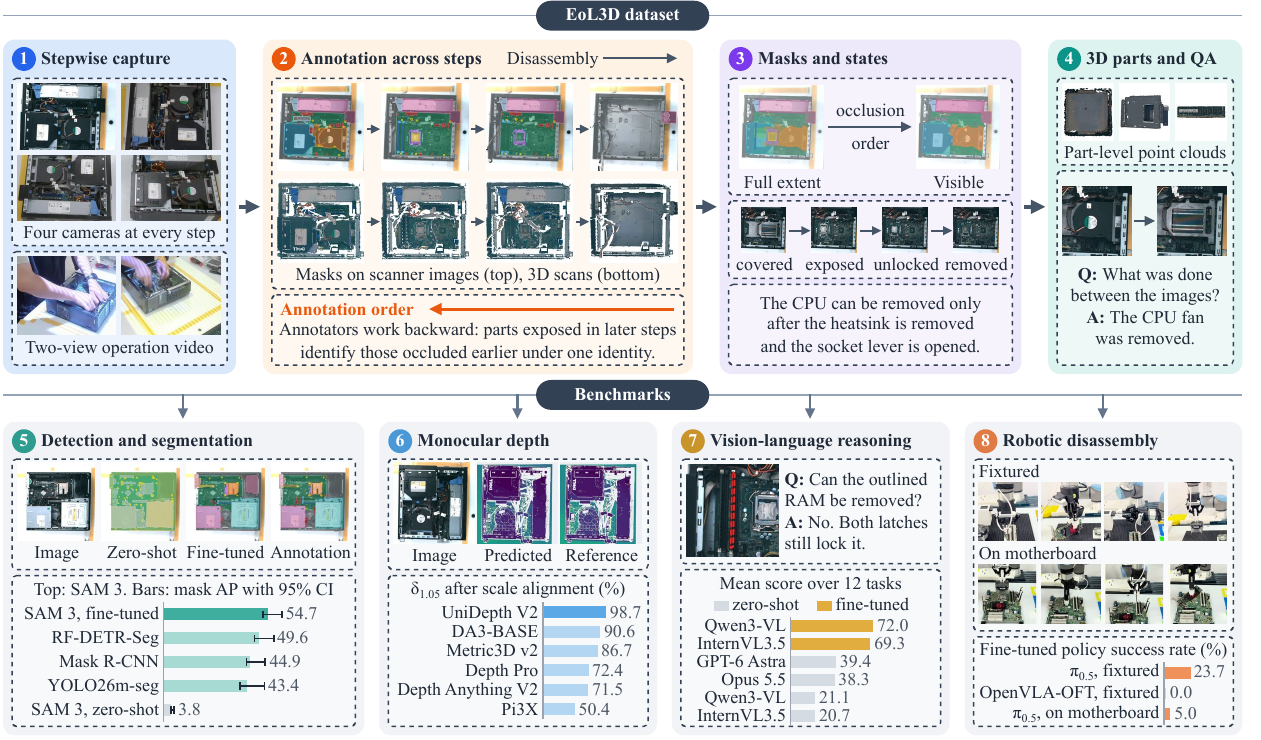}
    \caption{\textbf{From data collection to benchmarks in \dataset.} Human disassembly of EoL desktops is captured step by step from multiple views. Components are then annotated across steps under persistent instance identities with their masks and states, from which part-level point clouds and QA pairs are derived. The resulting data support benchmarks on component detection and segmentation, monocular depth estimation, vision-language reasoning, and robotic disassembly. Vision-language scores are averaged over the tasks in Table~\ref{tab:level2}.}
    \label{fig:pipeline}
\end{figure*}

\dataset records the desktop disassembly process through stepwise multi-view images, 3D reconstructions, human operation videos, and operation logs. Component annotations connect these observations across steps and provide the basis for part-level point clouds and QA pairs. A separate collection records robot demonstrations of component removal. The collection workflow is shown in Fig.~\ref{fig:overview}. We also summarize the data distribution and statistics in Fig.~\ref{fig:distributions}. Fig.~\ref{fig:pipeline} outlines how the collected data are built into the dataset and used in the benchmarks.

\subsection{Desktop Disassembly and Data Collection}

We collect disassembly sequences from 66 EoL desktops across 32 model families with different configurations and conditions within and across families. All desktops are disassembled at the same workbench location under the same observation setup, including three RGB-D cameras (two OAK-D Pro cameras and one Intel RealSense D435i), an overhead 3D scanner with RGB imaging, and two side-view video cameras for recording human operation. Each sequence begins with an initial observation and recording of the desktop type and condition; then the desktop case is secured to the workbench. The operator determines the next target, describes the operation, selects the proper tools, and completes the disassembly operation. Operations include disconnecting cables, releasing latches, unscrewing fasteners, and removing components. The target, operation description, and tools are recorded in the operation log.

The two video cameras record the human disassembly operations. The video segments are labeled with the descriptions provided by the operator. After the operator finishes each disassembly action, the scanner and three RGB-D cameras are used to record the remaining desktop assembly on the table. The disassembly process is considered finished when no more removable components remain inside the desktop case; then all dismantled components are placed on the workbench for component-level recording. The whole dataset contains 2,748 disassembly steps excluding initial observations, with 11,221 images across the four views. Images and operation logs are indexed by desktop and step. Acquisition specifications and the desktop inventory are provided in the appendix.

\subsection{Component Annotation}

We initially annotate with Label Studio~\cite{LabelStudio}, then develop a custom tool for cross-step component labeling, state annotation, and review. Two annotators manually label component-level instance segmentation across 2,814 scanner-view RGB images, including images of the removed components on the table. The annotations cover 3,217 parts across 24 classes with 84,225 instance annotations across images and corresponding bounding boxes, including structural parts, removable modules, fasteners, connectors, and latches. Annotators work backward through each disassembly sequence, using components that are exposed in later observations to identify those occluded in earlier images. Each component is assigned an instance ID that remains consistent across steps.

For each image, annotators draw the full-extent masks of the components and record which components occlude others. The masks and occlusion order are used to obtain the visible-region masks. When a component moves or changes state, its full-extent mask is redrawn for the affected frames. Regions that are never fully exposed during the sequence are completed manually. This process produces paired visible-region and full-extent masks for each annotated instance. State annotation follows the operation logs to record how components changed after each operation. The labels describe installation, removal, displacement, opening, and connection status. They also indicate whether a component still constrains the removal of another. For example, a latch can remain attached to the assembly after being released, while no longer securing the component it locks. These changes are recorded under the same instance ID throughout the sequence.

\subsection{3D Reconstruction and Part-Level Point Clouds}

We obtain 3D surface reconstructions of the remaining assembly at each step using fringe projection profilometry (FPP), a structured-light technique that projects fringe patterns onto the scene and reconstructs surface geometry from the observed phase using a calibrated imaging system~\cite{xu2020status, zhao2025precision}. Each reconstruction is registered to the corresponding scanner-view RGB image to obtain a colored point cloud of the remaining assembly. For each component, we use the visible-region mask to extract the corresponding surface points to form the part-level point cloud. After filtering by component class and reconstruction coverage, the dataset provides 31,705 part-level point clouds. We also project the reconstructed 3D points onto the corresponding image plane using the calibrated camera intrinsics. The resulting metric depth maps provide references for evaluating monocular depth estimation models.

\subsection{Disassembly Graphs and QA Generation}

We first establish a common set of disassembly rules that specify the relationships and prerequisites for releasing and removing components. For each desktop, these rules are applied to its annotated components and the relations present in its configuration. The components form the graph nodes, with 3,512 edges across the dataset describing fastening, connection, locking, coverage, and blocking relations. An independent researcher checks the segmentation masks and verifies each desktop's component relations. At each step, the graph and current component states determine the feasible operations and the prerequisites for each target. 

The disassembly graphs, images, annotations, and operation logs are used with predefined templates to construct QA pairs. Perception questions ask models to count or locate components, identify their states, and interpret observations from different views. Understanding questions use the operation logs and state timelines to ask about completed operations, changes over one or more steps, and disassembly progress. Planning questions additionally consider occlusion relations and rules for removing loose parts. They ask whether an operation is feasible, which operation can follow, or how subsequent operations should be ordered. Other questions ask models to check a proposed plan against the removal prerequisites. Each QA pair includes an answer linked to its source annotations or operation records, together with a step-by-step reasoning trace. Using these templates and source records, Codex from OpenAI first generates 300 QA pairs covering all question types. Two researchers manually review all questions and answers and provide comments for two additional rounds of revision to ensure quality. The revised templates are used to generate 184,263 QA pairs. A second Codex agent checks the question-generation process and the quality of the QA pairs. Another 300 QA pairs, distinct from the initial set, are sampled for manual review by two researchers. No errors are identified in this process.

\subsection{Robot Demonstrations of Component Removal}
\begin{table*}[t]
\caption{\textbf{Component detection and visible-region instance segmentation on the test desktops.} Brackets give 95\% confidence intervals obtained by resampling the test desktops. AP$_S$ and AP$_L$ are detection AP on small and large components. AP$_{\mathrm{vis}}$ and AP$_{\mathrm{occ}}$ are mask AP on unoccluded instances and on instances with at least 60\% of their extent hidden. The two frozen-SAM rows share the RF-DETR-Seg detection results. Bold marks the best result in each column.}
\label{tab:level1}
\centering\footnotesize
\setlength{\tabcolsep}{2pt}
\begin{tabular*}{\textwidth}{@{}l@{\extracolsep{\fill}}rrrrrrrrrrrr@{}}
\toprule
& \multicolumn{5}{c}{Detection} & \multicolumn{7}{c}{Instance segmentation} \\
\cmidrule(lr){2-6}\cmidrule(l){7-13}
Method & AP [95\% CI] & AP$_{50}$ & F1 & AP$_S$ & AP$_L$ & AP [95\% CI] & AP$_{50}$ & F1 & \shortstack[r]{Class\\mIoU} & \shortstack[r]{Pixel\\IoU} & AP$_{\mathrm{vis}}$ & AP$_{\mathrm{occ}}$ \\
\midrule
YOLO26m-seg & 45.0 \textnormal{[41.9, 55.0]} & 62.9 & 76.0 & 14.6 & 53.3 & 43.4 \textnormal{[39.9, 52.8]} & 62.0 & \textbf{75.8} & 49.7 & 64.6 & 45.1 & 6.7 \\
Mask R-CNN R50-FPN v2 & 51.8 \textnormal{[49.3, 58.9]} & 71.6 & 76.4 & 13.4 & 63.3 & 44.9 \textnormal{[43.1, 53.0]} & 64.3 & 75.0 & 51.3 & 76.0 & 45.6 & 5.9 \\
DINOv3 + Mask R-CNN & 42.9 \textnormal{[40.1, 52.0]} & 65.7 & 73.3 & 12.6 & 52.5 & 38.2 \textnormal{[35.7, 45.6]} & 63.1 & 72.4 & 47.7 & 70.4 & 37.2 & 3.9 \\
\midrule
RF-DETR-Seg & \multirow[c]{3}{*}{56.4 \textnormal{[52.1, 63.7]}} & \multirow[c]{3}{*}{72.2} & \multirow[c]{3}{*}{\textbf{77.2}} & \multirow[c]{3}{*}{14.9} & \multirow[c]{3}{*}{69.9} & 49.6 \textnormal{[47.5, 57.7]} & 70.8 & 75.2 & 57.3 & \textbf{83.1} & 49.0 & 11.6 \\
RF-DETR + frozen SAM 2.1 &  &  &  &  &  & 34.6 \textnormal{[33.8, 40.5]} & 56.7 & 62.2 & 49.5 & 54.7 & 34.3 & 1.4 \\
RF-DETR + frozen SAM 3 &  &  &  &  &  & 35.0 \textnormal{[33.6, 41.1]} & 57.9 & 64.6 & 49.5 & 54.7 & 36.3 & 1.4 \\
\midrule
SAM 3, zero-shot & 4.8 \textnormal{[4.5, 6.2]} & 5.7 & 7.0 & 0.4 & 6.0 & 3.8 \textnormal{[3.5, 5.2]} & 5.7 & 6.7 & 3.4 & 10.7 & 0.3 & 0.0 \\
SAM 3, fine-tuned & \textbf{59.3} \textnormal{[56.8, 66.2]} & \textbf{75.3} & 76.2 & \textbf{16.1} & \textbf{72.4} & \textbf{54.7} \textnormal{[51.9, 61.9]} & \textbf{73.7} & 75.6 & \textbf{61.7} & 79.0 & \textbf{56.8} & \textbf{12.1} \\
\bottomrule
\end{tabular*}
\end{table*}

Robot demonstrations are collected in two settings through a robotic disassembly system using a UR5e and a Robotiq 2F-85 gripper with customized extension fingers for disassembly. In the fixture setting, component slots and holders cut from motherboards are mounted on LEGO-type blocks that attach to a common base plate. In the on-motherboard setting, the target components remain installed on a motherboard removed from the desktop. Across the collection, targets include CPUs, CPU fans, RAM modules, connectors, graphics cards, hard disk drives, and solid-state drives. For each demonstration, an operator teleoperates the robot with dual joysticks to disassemble the target component and place it in the designated sorting zone. Each demonstration is paired with a language instruction. EoL3D also includes human correction data for any failed operation. Throughout the episode, wrist and base cameras record synchronized images at $256\times256$ resolution alongside robot states and commanded actions at 30 Hz. The collection contains 1,808 demonstrations, comprising 1,092 episodes in the fixture setting and 716 in the on-motherboard setting.

\begin{table*}[t]
\caption{\textbf{Vision-language reasoning on the test desktops.} Per-task scores of zero-shot and fine-tuned models, given the images or the question text alone (No image). Tasks are grouped into perception, understanding, and planning.}
\label{tab:level2}
\centering\footnotesize
\setlength{\tabcolsep}{1.7pt}
\begin{tabular*}{\textwidth}{@{}ll@{\extracolsep{\fill}}*{11}{c}@{}}
\toprule
& & & & \multicolumn{5}{c}{Zero-shot} & \multicolumn{4}{c}{Fine-tuned} \\
\cmidrule(lr){5-9}\cmidrule(l){10-13}
& & & & \multicolumn{4}{c}{Image} & \multicolumn{1}{c}{No image} & \multicolumn{2}{c}{Image} & \multicolumn{2}{c}{No image} \\
\cmidrule(lr){5-8}\cmidrule(lr){9-9}\cmidrule(lr){10-11}\cmidrule(l){12-13}
Task & Metric & $N$ & Const. & Qwen & InternVL & GPT-6 Astra & Opus 5.5 & GPT-6 Astra & Qwen & InternVL & Qwen & InternVL \\
\midrule
\multicolumn{13}{@{}l}{\textbf{Perception}} \\
Visible component boxes & Box-set F1 & 55 & 8.8 & 2.5\rlap{$^\dagger$} & 0.6 & 9.8 & 8.6 & --- & 18.9 & 17.9 & --- & --- \\
Component grounding & Grounding score & 47 & 2.1 & 2.1 & 4.3 & 52.1 & 34.9\rlap{$^\dagger$} & --- & 51.1 & 51.1 & --- & --- \\
Same step across views & Accuracy & 37 & 51.4 & 43.2 & 48.6 & 78.4 & 70.3 & --- & 86.5 & 83.8 & --- & --- \\
Component state or screw count & State/count score & 54 & 27.8 & 35.2 & 38.5 & 71.9 & 66.3 & --- & 83.3 & 83.3 & --- & --- \\
\midrule
\multicolumn{13}{@{}l}{\textbf{Understanding}} \\
Observed operation & Verb + target acc. & 54 & 20.4 & 3.7 & 3.7\rlap{$^\dagger$} & 24.1 & 25.9 & --- & 74.1 & 74.1 & --- & --- \\
State change & Accuracy & 74 & 51.4 & 56.8 & 56.8 & 59.5 & 77.0 & --- & 95.9 & 89.2 & --- & --- \\
Completed operations & Set F1 & 58 & 13.9 & 5.2 & 5.2 & 28.3 & 20.8 & --- & 75.1 & 70.4 & --- & --- \\
\midrule
\multicolumn{13}{@{}l}{\textbf{Planning}} \\
Proposed action feasibility & Accuracy & 71 & 46.5 & 46.5\rlap{$^\dagger$} & 54.9 & 70.4 & 66.2 & 53.5 & 73.2 & 63.4 & 45.1 & 50.7 \\
Next feasible action categories & Set F1 & 52 & 46.5 & 3.3 & 2.3 & 8.4 & 10.5 & --- & 71.2 & 65.3 & --- & --- \\
Target removal plan & Replay success & 53 & 66.0 & 0.0 & 0.0 & 0.0 & 1.9 & 0.0 & 98.1 & 100.0 & 67.9 & 67.9 \\
Constrained removal plan & Replay success & 51 & 5.9 & 0.0 & 0.0 & 2.0 & 2.0 & 0.0 & 49.0 & 47.1 & 27.5 & 27.5 \\
Proposed plan validity & Accuracy & 57 & 50.9 & 54.4 & 33.3\rlap{$^\dagger$} & 68.4 & 75.4 & 63.2 & 87.7 & 86.0 & 63.2 & 78.9 \\
\bottomrule
\end{tabular*}
\end{table*}

\section{Evaluation}

We split the data by desktop for detection, instance segmentation, and vision-language reasoning. All images and questions from a desktop belong to the same split, with 48 desktops for training, 6 for validation, and 12 for testing. We also evaluate monocular depth predictions against scan-derived reference maps and test VLA policies on component-level disassembly tasks. Benchmark implementation details and evaluation protocols are provided in the appendix.

\subsection{Component Detection and Segmentation}

We evaluate component detection and visible-region instance segmentation on 492 scanner-view images containing 12,930 annotated instances from the test desktops. YOLO26m-seg~\cite{sapkota2025yolo26}, Mask R-CNN R50-FPN v2~\cite{he2017mask,lin2017feature}, and RF-DETR-Seg~\cite{robinson2026rf} are fine-tuned. The Mask R-CNN comparison also includes a frozen DINOv3 backbone~\cite{simeoni2025dinov3} with a trainable feature pyramid and prediction heads. For SAM~3~\cite{carion2026sam}, we compare zero-shot predictions with those after fine-tuning, using the same 24 component class names as text prompts. To compare segmentation with fixed detections, RF-DETR-Seg boxes are also passed to frozen SAM~2.1~\cite{ravi2025sam} and SAM~3. These predictions retain the original class labels and confidence scores.

We report COCO AP~\cite{lin2014microsoft}, AP$_{50}$, and micro-F1 at IoU 0.50 for detection and instance segmentation. Class mIoU and pixel IoU are also reported for segmentation. Table~\ref{tab:level1} shows that SAM~3 performs poorly with class-name prompts alone but achieves the highest detection and mask AP after fine-tuning. Using RF-DETR-Seg detections to prompt frozen SAM~2.1 or SAM~3 results in lower mask AP than retaining RF-DETR-Seg's own trained mask head. Small-component detection remains difficult even for fine-tuned SAM~3, with a large gap between AP$_S$ and AP$_L$ that is also present in the other models. Occlusion also limits the performance of all the models.

\subsection{Vision-Language Reasoning}

\begin{table*}[t]
\caption{\textbf{Zero-shot monocular depth estimation.} Left: results at each model's original output scale. Right: results after per-image median scale alignment. Arrows give the better direction, and bold marks the best result in each column.}
\label{tab:depth}
\centering\footnotesize
\setlength{\tabcolsep}{2pt}
\begin{tabular*}{\textwidth}{@{}ll@{\extracolsep{\fill}}rrrrrrrrr@{}}
\toprule
& & \multicolumn{5}{c}{No scale alignment} & \multicolumn{4}{c}{After median scale alignment} \\
\cmidrule(lr){3-7}\cmidrule(l){8-11}
Model & Output scale & MAE (m) $\downarrow$ & RMSE (m) $\downarrow$ & AbsRel $\downarrow$ & $\delta_{1.25}$ $\uparrow$ & \shortstack[r]{Scale ratio\\(ideal 1)} & MAE (m) $\downarrow$ & RMSE (m) $\downarrow$ & AbsRel $\downarrow$ & $\delta_{1.05}$ $\uparrow$ \\
\midrule
UniDepth V2~\cite{piccinelli2025unidepthv2} & Metric & \textbf{0.398} & \textbf{0.400} & \textbf{0.330} & \textbf{2.96\%} & \textbf{1.334} & \textbf{0.016} & \textbf{0.021} & \textbf{0.014} & \textbf{98.73\%} \\
Depth Anything V2~\cite{yang2024depth} & Metric & 0.736 & 0.748 & 0.609 & 0.00\% & 1.650 & 0.046 & 0.071 & 0.040 & 71.49\% \\
Depth Pro~\cite{bochkovskiy2025depth} & Metric & 0.980 & 0.989 & 0.811 & 0.00\% & 1.862 & 0.038 & 0.061 & 0.033 & 72.37\% \\
Metric3D v2~\cite{hu2024metric3d} & Metric & 2.073 & 2.078 & 1.718 & 0.00\% & 2.754 & 0.023 & 0.034 & 0.019 & 86.66\% \\
\midrule
Pi3X~\cite{wang2026pi} & Approximate metric & 1.214 & 1.218 & 1.010 & 0.00\% & 2.023 & 0.062 & 0.072 & 0.052 & 50.37\% \\
DA3-BASE~\cite{lin2025depth} & Relative & -- & -- & -- & -- & 0.762$^\dagger$ & 0.029 & 0.035 & 0.024 & 90.58\% \\
\bottomrule
\end{tabular*}
\par\smallskip
\parbox{\textwidth}{\footnotesize $^\dagger$Descriptive ratio only; DA3-BASE does not output metric depth.}
\end{table*}

We evaluate the capabilities of perception, understanding, and planning on 663 questions from the 12 test desktops. Qwen3-VL-8B-Instruct~\cite{bai2025qwen3}, InternVL3.5-8B~\cite{wang2025internvl3}, GPT-6 Astra~\cite{openai2026astra}, and Claude Opus 5.5~\cite{anthropic2026opus55} are evaluated zero-shot. Qwen and InternVL are also fine-tuned on questions from the training split. Questions use the current assembly image or paired observations from consecutive steps or different cameras. Image inputs and their order are identical across models. For selected planning tasks, GPT-6 Astra and the fine-tuned models also receive the same question text without images or action history.

Table~\ref{tab:level2} reports scores averaged over each task's $N$ test questions using the listed \emph{Metric}. All scores are reported on a scale from 0 to 100. The \emph{Const.} baseline selects the most frequent complete reference answer in each task's test set. It repeats this answer for every question and is evaluated using the same metric. Operation recognition requires both the action verb and target to be correct. Plan replay checks predicted operations against component states and removal prerequisites. Any ordering that completes the required operations and reaches the goal under the stated constraints is accepted. Responses that do not follow the required answer format receive zero. Results are marked with $^\dagger$ when this occurs in at least 10\% of responses. Dashes indicate tasks cannot be evaluated without images.

From the results, the model fine-tuning improves component grounding, which requires locating a component in the image from its textual description. It also improves the recognition of operations and state changes, while predicting all visible component boxes remains difficult. Fine-tuning also helps with generating valid disassembly plans and substantially improves plan replay success, but success rates are lower when questions require certain operations to follow a specified order. Removing the images lowers the scores of both fine-tuned models on all four planning tasks tested without them.

\subsection{Monocular Depth Estimation}

We evaluate monocular depth using the initial scanner-view images from five desktops and the corresponding scan-derived reference maps. Depth Anything V2~\cite{yang2024depth}, Depth Pro~\cite{bochkovskiy2025depth}, UniDepth V2~\cite{piccinelli2025unidepthv2}, and Metric3D v2~\cite{hu2024metric3d} provide metric depth predictions. Pi3X~\cite{wang2026pi} provides approximate metric depth, while Depth Anything~3 (DA3-BASE)~\cite{lin2025depth} predicts relative depth. All models are evaluated without fine-tuning on \dataset.

The reference maps cover approximately 47 to 51\% of each image. Metrics are computed at pixels with valid predicted and reference depth, then averaged across desktops with equal weight. At the original scale, metric and approximate metric predictions are evaluated using mean absolute error (MAE), root mean squared error (RMSE), absolute relative error (AbsRel), and threshold accuracy. Their median predicted-to-reference depth ratios summarize scale bias. Each predicted map is then rescaled using a single median-based factor computed from its valid predicted and reference depths. All models are evaluated after alignment using the three error metrics and threshold accuracy. For predicted depth $\hat d$ and reference depth $d$, threshold accuracy $\delta_t$ is the percentage of valid pixels satisfying $\max(\hat d/d,d/\hat d)<t$. We use a stricter threshold accuracy $\delta_{1.05}$ after scale alignment to better assess residual depth errors, compared with $\delta_{1.25}$ before alignment.

Table~\ref{tab:depth} shows that the metric models and Pi3X overestimate depth on average. Their MAE, RMSE, and AbsRel decrease substantially after per-image scale alignment. This reduction identifies absolute scale as a major source of error in these desktop observations. However, this alignment requires FPP reference depth at evaluation time and therefore does not demonstrate independent metric-scale recovery.

\subsection{Robotic Component Disassembly}

The robotic disassembly system needs to disassemble the component specified in the instruction and place it in the designated sorting zone. We fine-tune OpenVLA-OFT~\cite{kim2025fine} and $\pi_{0.5}$~\cite{intelligence2025pi_} on complete demonstrations from the fixture setting. A separate $\pi_{0.5}$ policy is trained on the on-motherboard demonstrations. Each policy controls the arm and gripper from approach through extraction and placement. The fixture evaluation covers seven component categories at five positions each, with ten rollouts per position and 350 rollouts per policy. Both policies use the same test positions, instructions, and home pose on the same robotic system. A trial fails after 150s or when the operator judges completion infeasible. In the on-motherboard setting, each component is tested at two positions with ten rollouts per position, giving 100 rollouts. OpenVLA-OFT completes none of the fixtured trials, while $\pi_{0.5}$ succeeds in 23.7\%. The motherboard policy succeeds in 5\% of trials. Reliable execution of the full disassembly and placement sequence remains difficult in both settings.

\section{Conclusion}

To our knowledge, EoL3D is the first multimodal dataset covering the complete disassembly processes of 66 EoL desktops with different types and usage conditions, together with benchmarks across computer vision, vision-language reasoning, and robotic manipulation. Its instance records connect visual and geometric observations with component states and removal prerequisites throughout desktop disassembly. Fine-tuning improves perception and planning, while accurate metric depth estimation and reliable robotic execution remain difficult. Segmentation annotations currently cover only the scanner view. Future work will extend these annotations to the remaining camera views and expand the robot demonstration collection for in-chassis disassembly. We also plan to combine the existing 3D scans with Gaussian splatting to reconstruct each disassembly state in simulation environments for robotic learning.

\section*{LLM Usage Statement}
OpenAI's ChatGPT assists with grammar checking and draft formulation. OpenAI's Codex assists with QA pair building, coding, and benchmark evaluation. Claude Code assisted with software build coding, image structuring, and dataset structure build.

\bibliographystyle{IEEEtran}
\bibliography{ref}

\clearpage
\appendices

\section{Data Collection and Desktop Inventory}
\label{sec:app_collection}

\subsection{Setup and Inventory}
\label{sec:app_setup}
\label{sec:app_desktops}

All desktops are disassembled at the same workbench under a fixed observation setup (\cref{tab:suppl_hardware}). A custom fringe-projection structured-light scanner looks down at the desktop. From the images captured during one fringe-projection acquisition, it reconstructs the surface and recovers an RGB color image. We use this RGB image to define the Scanner view. Two OAK-D Pro cameras provide the near-top-down OAK-D top view and the oblique OAK-D side view. A RealSense D435i provides the RealSense view. These three RGB-D cameras also record depth aligned with their color images. The four views record the same desktop state after each disassembly step. Two side-view video cameras record the human operations.

\begin{table*}[!tp]
\caption{\textbf{Sensors and recorded data.} The wrist and base streams belong to separate robot demonstrations.}
\label{tab:suppl_hardware}
\centering
\small
\setlength{\tabcolsep}{4pt}
\begin{tabular}{@{}p{0.17\textwidth}p{0.30\textwidth}p{0.47\textwidth}@{}}
\toprule
Stream & Sensor & Recorded data \\
\midrule
Scanner view & Custom fringe-projection structured-light scanner, top-down & $1600\times1600$ RGB image and 3D reconstruction, both recovered from one fringe-projection capture \\
OAK-D top view & Luxonis OAK-D Pro, near top-down & $4032\times3040$ RGB image, aligned $1280\times800$ RGB-D, and point cloud \\
OAK-D side view & Luxonis OAK-D Pro, oblique & Same as the OAK-D top view \\
RealSense view & Intel RealSense D435i & $1280\times720$ aligned RGB-D and point cloud \\
Human operation video & Two side-view video cameras & 3,179 paired clips, 11.1~h, 30 frames per second, faces masked \\
\midrule
Robot wrist view & Intel RealSense D435i on the gripper & $256\times256$ RGB at 30~Hz \\
Robot base view & Luxonis OAK-D Pro, fixed & $256\times256$ RGB at 30~Hz \\
\bottomrule
\end{tabular}
\end{table*}

Each desktop is a distinct physical unit. Desktops of the same model can differ in installed components, cabling, and usage condition, as shown in \cref{tab:app_desktops}. The data are split by desktop into 48 training, 6 validation, and 12 test desktops.

\begin{table*}[!tp]
\caption{\textbf{Desktop inventory and configuration.} Form: S = slim chassis, T = tower. Split: Tr = training, Va = validation, Te = test. Steps exclude the initial observation. Parts are the annotated instances of the desktop. RAM, Drv., Card, Conn., and Scr. count its memory modules, storage and optical drives, expansion cards, cable connectors, and screws.}
\label{tab:app_desktops}
\centering
\scriptsize
\setlength{\tabcolsep}{2pt}
\begin{tabular}{@{}llccrrrrrrr@{\hspace{7pt}}llccrrrrrrr@{}}
\toprule
ID & Model & Form & Split & Steps & Parts & RAM & Drv. & Card & Conn. & Scr. & ID & Model & Form & Split & Steps & Parts & RAM & Drv. & Card & Conn. & Scr. \\
\midrule
D01 & HP EliteDesk 800 G2 & T & Tr & 36 & 43 & 2 & 2 & 0 & 13 & 12 & D34 & HP Pro 3500 & T & Tr & 36 & 38 & 1 & 2 & 0 & 13 & 12 \\
D02 & Dell OptiPlex 7020 & S & Tr & 47 & 50 & 2 & 2 & 0 & 17 & 10 & D35 & Dell OptiPlex 380 & S & Va & 41 & 40 & 2 & 2 & 1 & 12 & 11 \\
D03 & Dell OptiPlex 7010 & S & Va & 47 & 52 & 2 & 2 & 1 & 13 & 16 & D36 & Dell (unrecorded) & S & Tr & 41 & 45 & 2 & 3 & 0 & 13 & 11 \\
D04 & Dell OptiPlex 3020 & S & Te & 45 & 47 & 2 & 2 & 1 & 17 & 10 & D37 & Dell OptiPlex 7020 & T & Tr & 44 & 49 & 2 & 2 & 0 & 17 & 14 \\
D05 & Dell OptiPlex 7020 & S & Tr & 45 & 50 & 2 & 2 & 0 & 17 & 10 & D38 & Dell OptiPlex 7020 & T & Tr & 36 & 45 & 0 & 2 & 0 & 17 & 12 \\
D06 & Dell OptiPlex 7020 & S & Tr & 48 & 51 & 3 & 2 & 0 & 17 & 10 & D39 & Dell (unrecorded) & T & Tr & 53 & 55 & 4 & 2 & 3 & 18 & 12 \\
D07 & Dell OptiPlex 7020 & S & Tr & 55 & 61 & 4 & 2 & 0 & 15 & 16 & D40 & Dell OptiPlex 7010 & T & Tr & 36 & 44 & 1 & 2 & 0 & 15 & 12 \\
D08 & Dell OptiPlex 7010 & S & Tr & 45 & 57 & 1 & 2 & 0 & 14 & 16 & D41 & Dell OptiPlex 7010 & T & Va & 34 & 44 & 0 & 2 & 0 & 16 & 12 \\
D09 & Dell OptiPlex 7020 & S & Tr & 45 & 50 & 2 & 2 & 0 & 17 & 10 & D42 & HP EliteDesk 800 G1 & T & Tr & 41 & 48 & 1 & 2 & 1 & 17 & 12 \\
D10 & Dell OptiPlex 3020 & S & Te & 41 & 45 & 1 & 2 & 0 & 17 & 10 & D43 & Dell OptiPlex 9020 & T & Tr & 46 & 51 & 2 & 2 & 1 & 19 & 12 \\
D11 & Dell OptiPlex 7010 & S & Tr & 46 & 58 & 1 & 2 & 0 & 15 & 16 & D44 & HP EliteDesk 800 G1 & T & Tr & 38 & 47 & 0 & 2 & 1 & 17 & 12 \\
D12 & Dell OptiPlex 7020 & S & Tr & 45 & 50 & 2 & 2 & 0 & 17 & 10 & D45 & Custom-built & T & Tr & 35 & 41 & 2 & 1 & 1 & 14 & 7 \\
D13 & Dell OptiPlex 7020 & S & Tr & 41 & 46 & 2 & 1 & 0 & 14 & 10 & D46 & HP Z440 & T & Te & 35 & 50 & 1 & 1 & 1 & 11 & 14 \\
D14 & Dell OptiPlex 3010 & S & Tr & 47 & 55 & 1 & 2 & 0 & 16 & 16 & D47 & HP EliteDesk 800 G1 & T & Tr & 41 & 48 & 1 & 2 & 1 & 17 & 12 \\
D15 & Dell OptiPlex 3020 & S & Te & 41 & 45 & 1 & 2 & 0 & 17 & 10 & D48 & Lenovo M91p & S & Te & 35 & 44 & 1 & 2 & 0 & 12 & 11 \\
D16 & Dell OptiPlex 3010 & S & Tr & 44 & 53 & 1 & 2 & 0 & 14 & 16 & D49 & Dell Precision T5500 & T & Te & 32 & 44 & 4 & 1 & 4 & 11 & 4 \\
D17 & Dell OptiPlex 7020 & S & Tr & 45 & 50 & 2 & 2 & 0 & 17 & 10 & D50 & HP EliteDesk 800 G1 & T & Tr & 42 & 49 & 1 & 2 & 1 & 17 & 12 \\
D18 & Dell OptiPlex 7010 & S & Tr & 47 & 58 & 2 & 2 & 0 & 14 & 16 & D51 & HP EliteDesk 800 G1 & T & Tr & 40 & 47 & 1 & 2 & 0 & 17 & 12 \\
D19 & Dell OptiPlex 990 & S & Tr & 54 & 60 & 4 & 2 & 0 & 14 & 16 & D52 & HP Compaq 8200 Elite & T & Tr & 40 & 50 & 1 & 2 & 0 & 18 & 14 \\
D20 & Dell OptiPlex 3020 & S & Te & 45 & 46 & 2 & 2 & 0 & 17 & 10 & D53 & HP Compaq 8200 Elite & T & Tr & 47 & 55 & 4 & 2 & 1 & 18 & 14 \\
D21 & Dell OptiPlex 7020 & S & Va & 42 & 49 & 1 & 2 & 0 & 17 & 10 & D54 & Dell (unrecorded) & T & Tr & 45 & 51 & 2 & 2 & 1 & 17 & 12 \\
D22 & HP Compaq 8200 Elite & S & Tr & 43 & 47 & 2 & 2 & 0 & 14 & 12 & D55 & HP Compaq 8200 Elite & T & Tr & 47 & 55 & 4 & 2 & 1 & 18 & 14 \\
D23 & HP Compaq 8100 Elite & S & Tr & 41 & 47 & 2 & 2 & 0 & 14 & 12 & D56 & HP Z230 & T & Tr & 45 & 52 & 4 & 2 & 1 & 15 & 14 \\
D24 & HP Compaq 8300 Elite & S & Te & 41 & 49 & 1 & 2 & 1 & 16 & 12 & D57 & HP Compaq 8200 Elite & T & Tr & 46 & 54 & 4 & 2 & 1 & 17 & 14 \\
D25 & HP Compaq Pro 4300 & S & Tr & 40 & 44 & 1 & 2 & 0 & 16 & 12 & D58 & HP EliteDesk 800 G2 & T & Tr & 45 & 50 & 2 & 2 & 1 & 17 & 12 \\
D26 & HP Compaq 8100 Elite & S & Tr & 37 & 47 & 0 & 2 & 0 & 16 & 12 & D59 & HP Compaq 8200 Elite & T & Tr & 43 & 51 & 4 & 2 & 0 & 16 & 14 \\
D27 & HP ProDesk 400 G1 & S & Tr & 42 & 47 & 1 & 2 & 0 & 16 & 12 & D60 & Dell OptiPlex 9020 & T & Va & 45 & 51 & 2 & 2 & 1 & 17 & 12 \\
D28 & HP EliteDesk 800 G1 & S & Tr & 46 & 51 & 2 & 2 & 0 & 14 & 12 & D61 & HP Pavilion a6257c & T & Tr & 39 & 48 & 0 & 2 & 2 & 16 & 11 \\
D29 & HP Compaq 8200 Elite & S & Tr & 46 & 50 & 2 & 2 & 1 & 16 & 12 & D62 & Dell 5500 & T & Tr & 37 & 50 & 4 & 2 & 4 & 14 & 9 \\
D30 & HP Compaq Pro 4300 & S & Te & 35 & 37 & 1 & 2 & 0 & 9 & 12 & D63 & Dell Precision T7500 & T & Tr & 36 & 59 & 6 & 1 & 1 & 15 & 13 \\
D31 & HP Compaq Pro 4300 & S & Te & 42 & 44 & 1 & 2 & 0 & 16 & 12 & D64 & Apple Power Mac G4 & T & Tr & 21 & 24 & 1 & 1 & 1 & 5 & 4 \\
D32 & HP Compaq 8300 Elite & S & Te & 46 & 50 & 2 & 2 & 1 & 16 & 12 & D65 & Dell OptiPlex 3020 & T & Te & 42 & 51 & 2 & 2 & 1 & 17 & 12 \\
D33 & HP Compaq 8100 Elite & S & Va & 43 & 47 & 2 & 2 & 0 & 14 & 12 & D66 & HP Compaq 8200 Elite & T & Tr & 11 & 51 & 4 & 2 & 0 & 16 & 14 \\
\bottomrule
\end{tabular}
\end{table*}

\subsection{Sequence and Component Recordings}
\label{sec:app_capture}
\label{sec:app_video}
\label{sec:app_gallery}

Each disassembly sequence starts with an initial observation of the untouched desktop. After each disassembly step, an observation records the remaining assembly from the four views. We write S$k$ for the $k$-th observation of a desktop, so S01 is the initial observation. The operation log records the target component, operation, and tool for each disassembly step.

The two side-view video cameras record the operator throughout the disassembly. The recordings are divided into clips by operation. Each clip is labeled with the operator's description of that operation. The videos have no audio, and faces are masked. The data collection passed all required reviews, and the operator in the videos consented to the recording and to its release.

After a desktop is fully disassembled, all removed components are placed on the workbench and recorded with the same setup. These images are annotated separately from the disassembly sequences and used for component-level detection and segmentation.

\section{Component Annotation}
\label{sec:app_annotation}

\subsection{Classes, States, and Mask Scope}
\label{sec:app_classes}

The annotations cover 24 classes, each with its own states, as shown in \cref{tab:app_classes}. A connector is annotated when it is plugged into its socket. Cables are not annotated as parts in the disassembly sequences. Every RAM latch is annotated, including those at empty slots. An optical drive removed on its own is annotated as an optical drive. An optical drive that stays in its bracket throughout the sequence is annotated together with the bracket as a drive cage. The chassis is annotated in every Scanner image of a disassembly sequence.

\begin{table*}[!tp]
\caption{\textbf{Annotation classes and states.} Parts and Masks count the annotated instances of each class and their masks. States are listed in the order in which they usually occur.}
\label{tab:app_classes}
\centering
\footnotesize
\setlength{\tabcolsep}{4pt}
\begin{tabular}{@{}ll>{\raggedright\arraybackslash}p{0.38\textwidth}>{\raggedright\arraybackslash}p{0.23\textwidth}rr@{}}
\toprule
Class & Group & Definition & States & Parts & Masks \\
\midrule
chassis & Structural & Metal enclosure of the desktop; present in every Scanner image of a sequence & present & 66 & 2,800 \\
cover & Structural & Removable cover, air shroud, or duct & closed, open, removed & 64 & 1,809 \\
drive cage & Structural & Cage or bracket holding storage or optical drives & installed, displaced, removed & 55 & 1,609 \\
cooler bracket & Structural & Bracket that holds the CPU cooler down & installed, removed & 2 & 13 \\
screw hole & Structural & Hole left by a removed CPU cooler screw & present, removed & 35 & 510 \\
\addlinespace[2pt]
motherboard & Module & Main circuit board & installed, removed & 66 & 2,714 \\
cpu & Module & Processor in its socket & installed, removed & 65 & 1,854 \\
cpu cooler & Module & CPU heatsink, fan, or both & installed, displaced, removed & 75 & 783 \\
ram module & Module & Memory module & installed, removed & 127 & 3,225 \\
psu & Module & Power supply unit & installed, displaced, removed & 65 & 1,810 \\
storage drive & Module & Hard disk drive or solid-state drive & installed, displaced, removed & 65 & 752 \\
optical drive & Module & Optical drive that is taken out of the desktop & installed, removed & 29 & 279 \\
expansion card & Module & Add-in card in a slot, such as a graphics card & installed, removed & 36 & 1,060 \\
case fan & Module & Fan mounted on the case & installed, removed & 5 & 45 \\
case connector part & Module & Front I/O module of the case & installed, removed & 1 & 23 \\
\addlinespace[2pt]
screw & Fastener & Screw, with the part it fastens; a captive screw stays on its part & fastened, loosened, removed & 791 & 20,920 \\
\addlinespace[2pt]
connector & Connector & Cable plug in its socket; annotated while plugged in & plugged, unplugged, removed & 1,015 & 18,510 \\
\addlinespace[2pt]
ram latch & Latch & Latch at the end of a RAM slot, on occupied and empty slots & closed, open, removed & 494 & 20,245 \\
cpu socket lever & Latch & Retention lever and frame of the CPU socket & closed, open, removed & 65 & 2,694 \\
psu latch & Latch & Latch or bracket holding the power supply & closed, removed & 20 & 735 \\
drive latch & Latch & Latch holding a drive or a drive cage & closed, open, removed & 20 & 251 \\
card latch & Latch & Latch holding expansion cards & closed, open, removed & 16 & 564 \\
cooler latch & Latch & Latch, clip, or frame holding the CPU cooler & closed, removed & 5 & 6 \\
cable clip & Latch & Clip holding a cable & closed, open, removed & 35 & 1,014 \\
\bottomrule
\end{tabular}
\end{table*}

Every part has a state at each disassembly step. State changes follow the operations recorded in the operation log. Unscrewing changes a captive screw to \texttt{loosened} and a non-captive screw to \texttt{removed}. Disconnecting changes a connector to \texttt{unplugged}. Removing a part also removes the parts attached to it (\cref{sec:app_precedence}).

In disassembly-sequence images, a part has masks until it is removed, and a connector has masks until it is unplugged. Opened covers, opened latches, displaced parts, and loosened captive screws keep their masks. A fully hidden part has a full-extent mask and an empty visible-region mask.

\subsection{Reverse-Order Annotation}
\label{sec:app_layers}

\begin{figure*}[!tp]
\centering
\includegraphics[width=\textwidth]{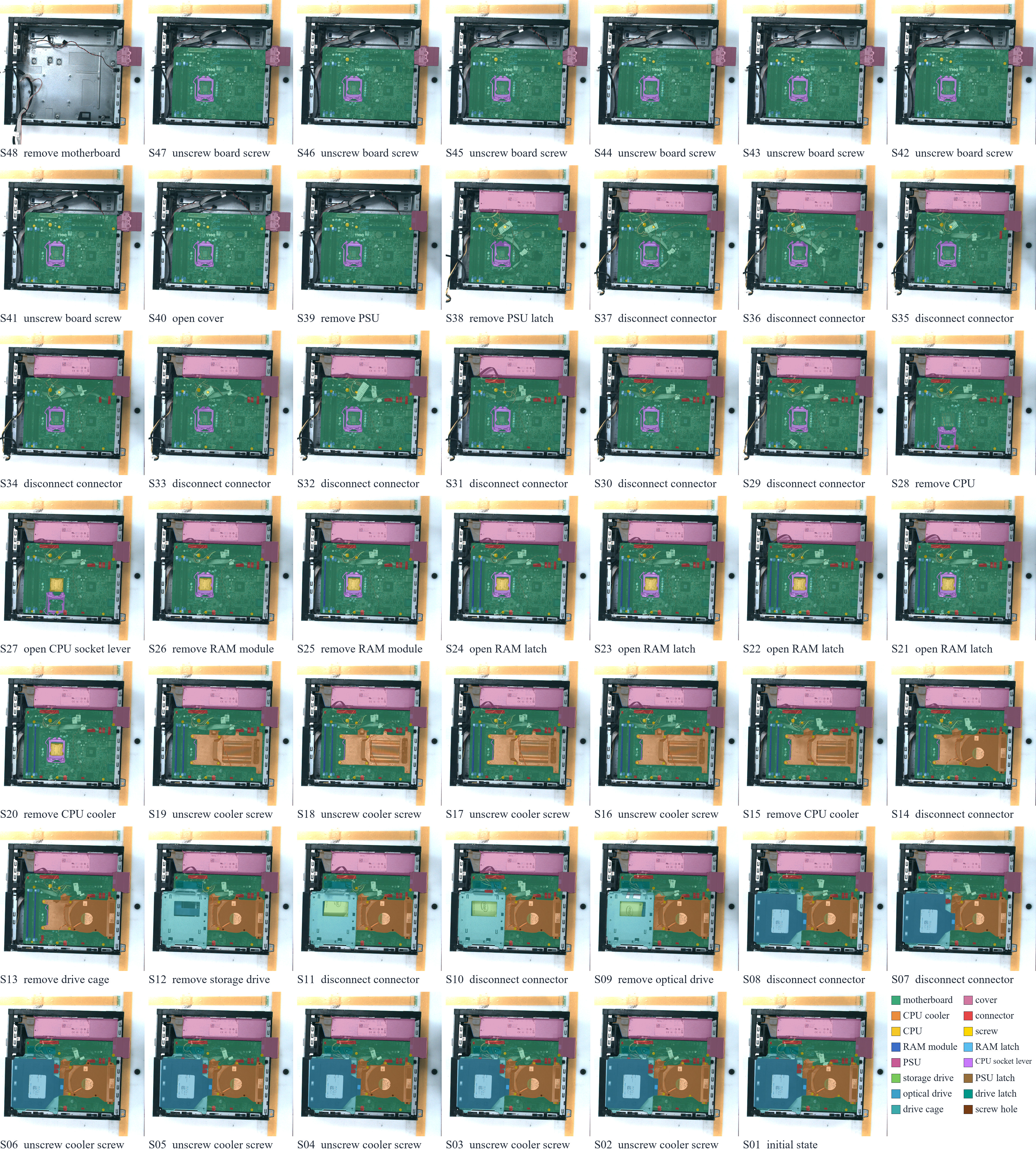}
\caption{\textbf{Complete annotated sequence of desktop D18.} Scanner images with visible-region masks run backward from S48 at the top left to the initial observation, S01. Except for S01, each label names the operation that produced the observation. Moving backward restores the configuration before each operation: a removed part returns, or a part returns to an earlier state. Colors indicate classes. The chassis is annotated but left uncolored.}
\label{fig:app_sequence}
\end{figure*}

Two annotators work backward through each sequence, from the last observation to the initial observation. Working backward serves two purposes. First, each part is drawn at the latest observation in which it is still in place, where parts that cover it in earlier observations have already been removed. This gives an accurate full-extent mask for a part that is hidden in earlier disassembly steps. The mask is reused in earlier observations while the part and the chassis remain unchanged, and it is redrawn when the part moves, opens, or is displaced, or when the chassis is reoriented. Second, the parts return one at a time, which makes the occlusion relations between parts clear. Each part is assigned a layer representing physical stacking, with the chassis below the motherboard, and the layer order records which parts occlude others (\cref{fig:app_sequence}).

\begin{figure*}[!tp]
\centering
\includegraphics[width=\textwidth]{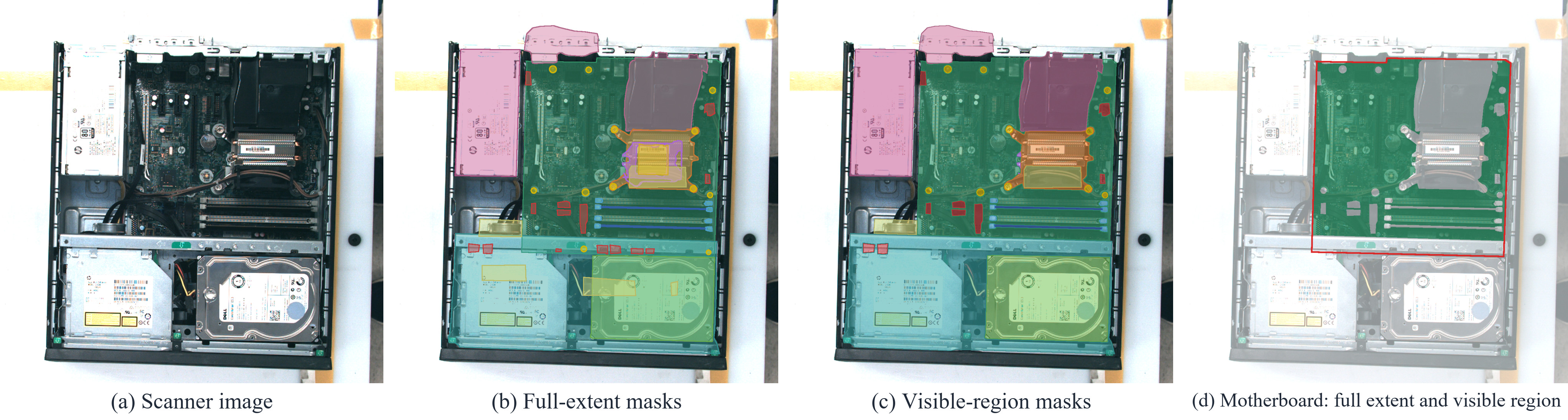}
\caption{\textbf{Full-extent and visible-region masks on one Scanner image (D28, initial observation).} (a) The image. (b) Full-extent masks of all parts. (c) Visible-region masks. (d) The motherboard, with its full extent outlined in red and its visible region filled. Colors as in \cref{fig:app_sequence}.}
\label{fig:app_masks}
\end{figure*}

A visible-region mask is obtained by subtracting the masks of higher-layer parts from the full-extent mask (\cref{fig:app_masks}). Annotators correct the layer order of a pair of parts when needed. They also correct individual visible-region masks where cables or unlabeled objects cover a part.

\begin{table*}[!tp]
\caption{\textbf{Relation types of the disassembly graphs.} An edge blocks the listed operations on a part until its release condition holds.}
\label{tab:app_relations}
\centering
\small
\setlength{\tabcolsep}{5pt}
\begin{tabular}{@{}l>{\raggedright\arraybackslash}p{0.36\textwidth}>{\raggedright\arraybackslash}p{0.17\textwidth}>{\raggedright\arraybackslash}p{0.30\textwidth}@{}}
\toprule
Relation & Meaning & Blocked operations & Released when \\
\midrule
Connection & A cable connector is plugged into the part or belongs to its cable & remove & The connector is unplugged \\
Fastening & A screw fastens the part & remove, displace, open & The screw is loosened or removed \\
Coverage & Another part lies over the part & all & The covering part is opened, displaced, or removed \\
Locking & A latch or lever locks the part & remove, displace, open & The latch is open \\
Blocking & A module is still mounted on the motherboard, or a drive is still in its cage & remove, displace & The mounted or contained part is removed \\
\bottomrule
\end{tabular}
\end{table*}

We first manually label 10 desktops and use these annotations to fine-tune an RF-DETR detector~\cite{robinson2026rf} and SAM~2.1~\cite{ravi2025sam}. The models identify what differs between the previous image and the current one, which reduces the labeling workload. The annotators then label each image manually, starting from the models' suggestions.

\subsection{Visibility and Review}
\label{sec:app_visibility}
\label{sec:app_qc}

The occlusion ratio of a part in an image is the hidden share of its full-extent area. The three occlusion levels in \cref{fig:distributions}(b) correspond to ratios below 30\%, from 30\% to below 95\%, and at least 95\%.

The annotations are reviewed in three stages. First, after a sequence is annotated, an annotator checks every observation again, including masks affected by later edits. Second, a reviewer examines pairs of large overlapping parts and decides whether the upper part blocks the lower part. Confirmed blocking pairs are recorded as coverage relations. Third, an independent researcher checks the masks and component relations of every desktop. Automatic checks then verify that the review is complete and that every visible-region mask lies within its full-extent mask.

\section{Relations and Feasible Operations}
\label{sec:app_precedence}

\begin{figure*}[!tp]
\centering
\includegraphics[width=0.72\textwidth]{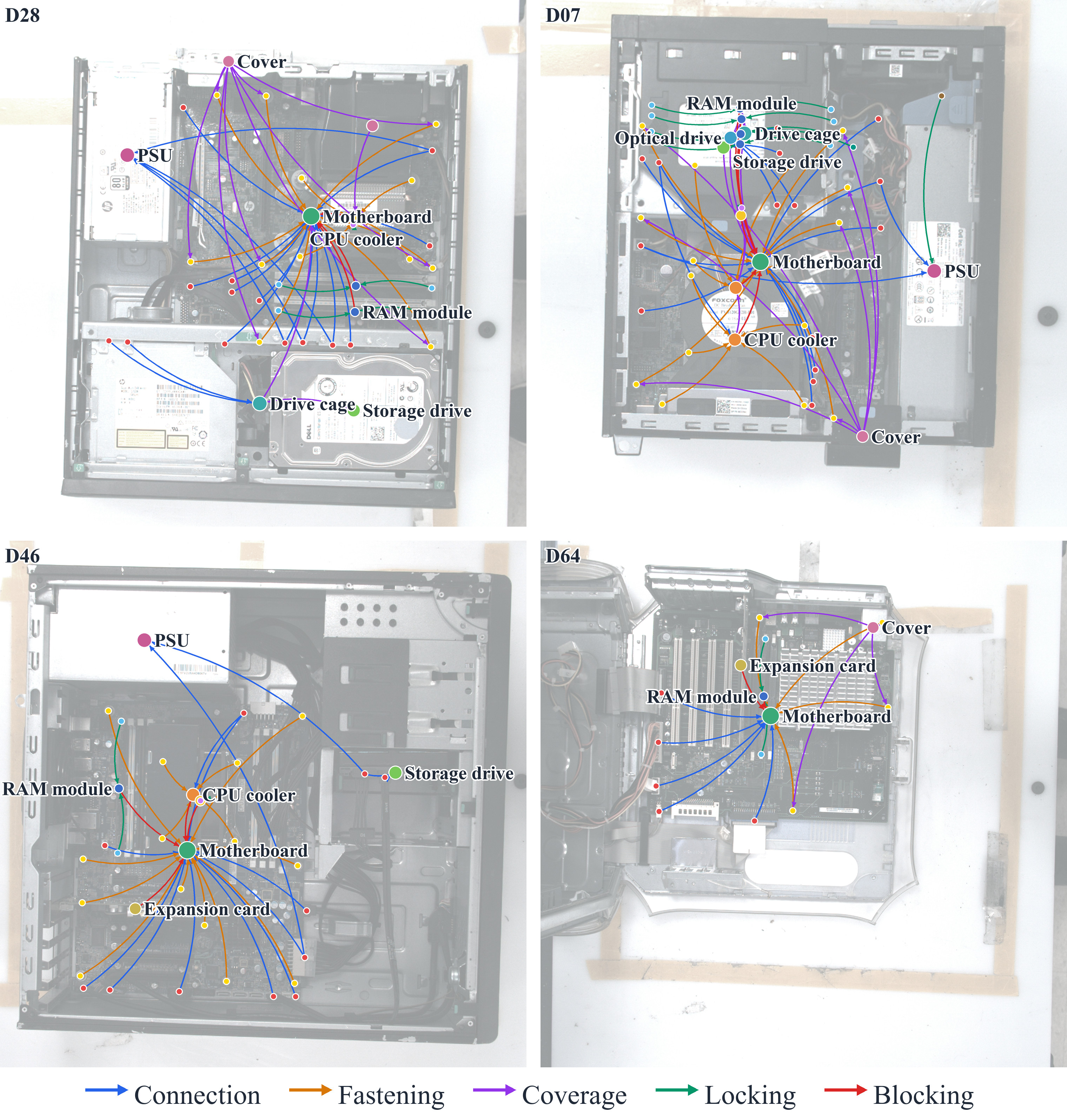}
\caption{\textbf{Disassembly graphs of four desktops of different models, drawn on their initial Scanner images.} The desktops are D28, D07, D46, and D64 of \cref{tab:app_desktops}. A node is an annotated part, placed at the center of its full-extent box and colored by class as in \cref{fig:app_sequence}. An arrow runs from a blocking part to the part it blocks. Its color gives the relation type of \cref{tab:app_relations}. The images are lightened for display.}
\label{fig:app_graph}
\end{figure*}

Each desktop has a disassembly graph that records which parts must be handled before others. We define each annotated part as a node, and the edges connecting the nodes define their relations. For each part, the annotation records what it is attached to or mounted on and which fasteners, connectors, and latches hold it. These records, the disassembly rules, and the coverage relations define the edges. \Cref{fig:app_graph} shows the graphs of four desktops.

\subsection{Component Relations and Disassembly Rules}

\Cref{tab:app_relations} lists the five relation types, the operations each one blocks, and the condition that releases it. Attached parts are removed with the part they are attached to, without a separate removal operation. RAM latches, CPU socket levers, and a card latch are removed with the motherboard. Captive CPU cooler screws are removed with the CPU cooler. Drive latches are removed with the drive cage. Other related parts must be handled before the part they block.

The disassembly rules state prerequisites that hold for every desktop. A CPU cooler can be removed once its screws are loosened or removed and its fan connector is unplugged. An air shroud must be opened or removed first. The cooler is removed before the CPU socket lever is opened. The lever is opened before the CPU is removed. Both latches of a RAM slot must be open before the module is removed. A card latch must be open before an expansion card is removed. The motherboard can be removed once its screws and mounted modules are removed and its connectors are unplugged.

Coverage relations come from three sources: the disassembly rules, the reviewer decisions in \cref{sec:app_qc}, and mask overlap for selected pairs of classes. The overlap test compares the upper part's visible-region mask with the lower part's full-extent mask. A drive cage that lies over other parts must be displaced or removed first.

\subsection{Feasibility at a Step}
\label{sec:suppl_graph}

Whether an operation is feasible at a disassembly step depends on four conditions. For assembly state $s$, operation $a$, and part $p$, let $I$ denote applicability, $R$ release of relations in the disassembly graph, $L$ priority of loose parts, and $V$ coverage in the current observation. An operation is feasible when all four hold:
\begin{equation}
\begin{split}
\mathcal{A}(s)=\{(a,p):\;&I(a,p,s)\land R(a,p,s)\\
&\land L(a,p,s)\land V(a,p,s)\}.
\end{split}
\label{eq:permitted_actions}
\end{equation}
Applicability requires that the operation is allowed for the part's class and current state and that the part has not been removed. The relation condition holds when every relation blocking the operation on the part has been released. A loose part is one that lay unattached inside the desktop from the start. While such a part is still present, only operations on loose parts or operations required for their removal are feasible.

The coverage condition $V$ is evaluated from the full-extent masks and layer order in the current Scanner image. An upper part that covers at least 20\% of a lower part's full-extent area must be handled first, unless it has been opened, displaced, unplugged, or removed. Cables, the chassis, screw holes, case fans that are not loose, and drive cages that never move during the sequence do not count as covering parts. Same-class pairs among connectors, RAM latches, RAM modules, and screws are also excluded. Plan generation and plan checking apply all four conditions at every step of a plan. Both use the covering pairs computed once from the Scanner image of the question's observation, and a pair stops blocking once an earlier plan step opens, displaces, unplugs, or removes its upper part.

\subsection{Consistency with Operation Logs}

Replaying each operation log against the graph and step conditions shows that every recorded operation is feasible at its disassembly step. Steps involving several parts are replayed one operation at a time.

\section{3D Reconstruction and Part-Level Point Clouds}
\label{sec:app_3d}

\begin{figure*}[!tp]
\centering
\includegraphics[width=\textwidth]{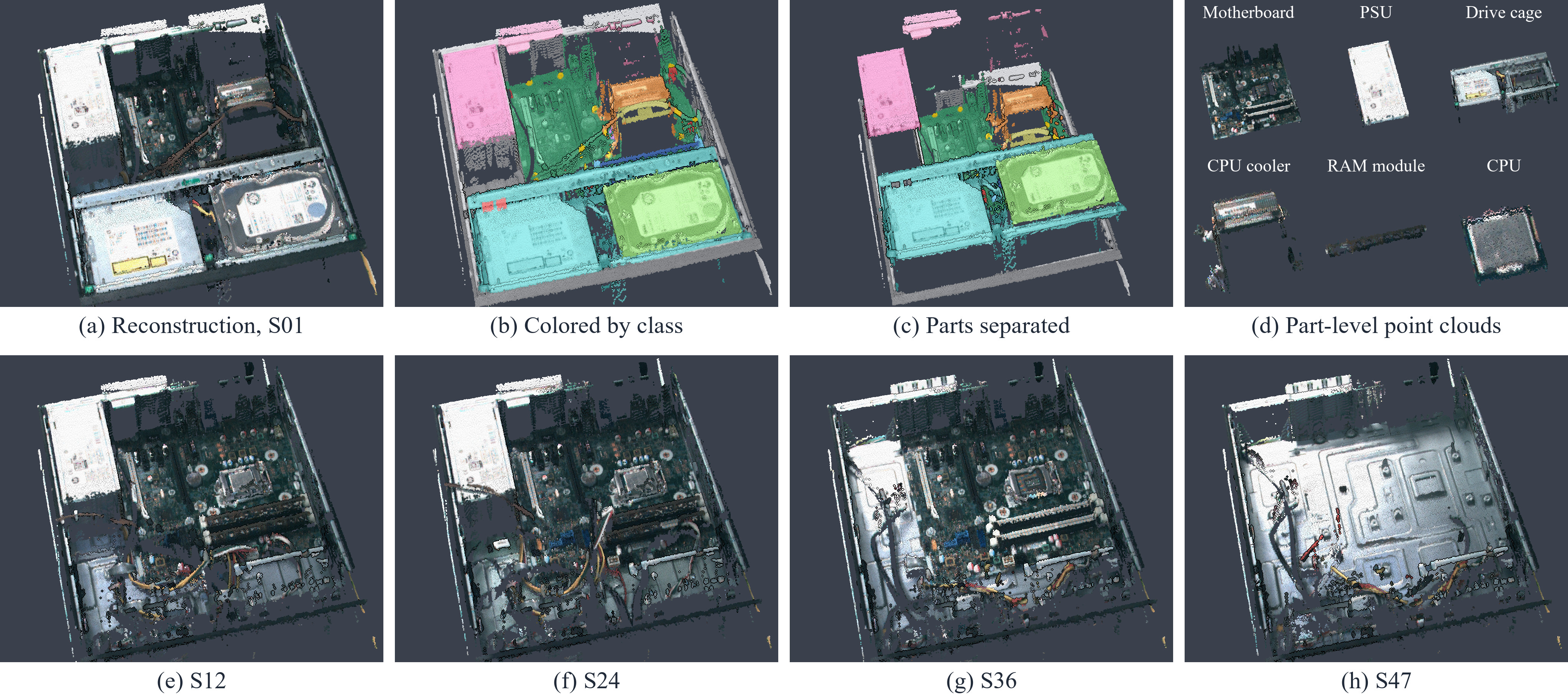}
\caption{\textbf{Reconstruction and part-level points of desktop D28.} Points are assigned to parts through the visible-region masks and shown before the filtering of the released part-level point clouds, so the chassis and the storage drive are included. (a) S01 seen obliquely. (b) The assigned points, colored by class as in \cref{fig:app_sequence}; the chassis is gray. (c) The same points displaced for display along the scanner viewing axis by a fixed distance per class, from 0 to 150~mm. (d) Part-level point clouds of six classes. For each class, the part and observation with the most reconstructed points are shown. Each part is scaled separately. (e)--(h) Later observations.}
\label{fig:app_pointcloud}
\end{figure*}

\begin{figure*}[!tp]
\centering
\includegraphics[width=\textwidth]{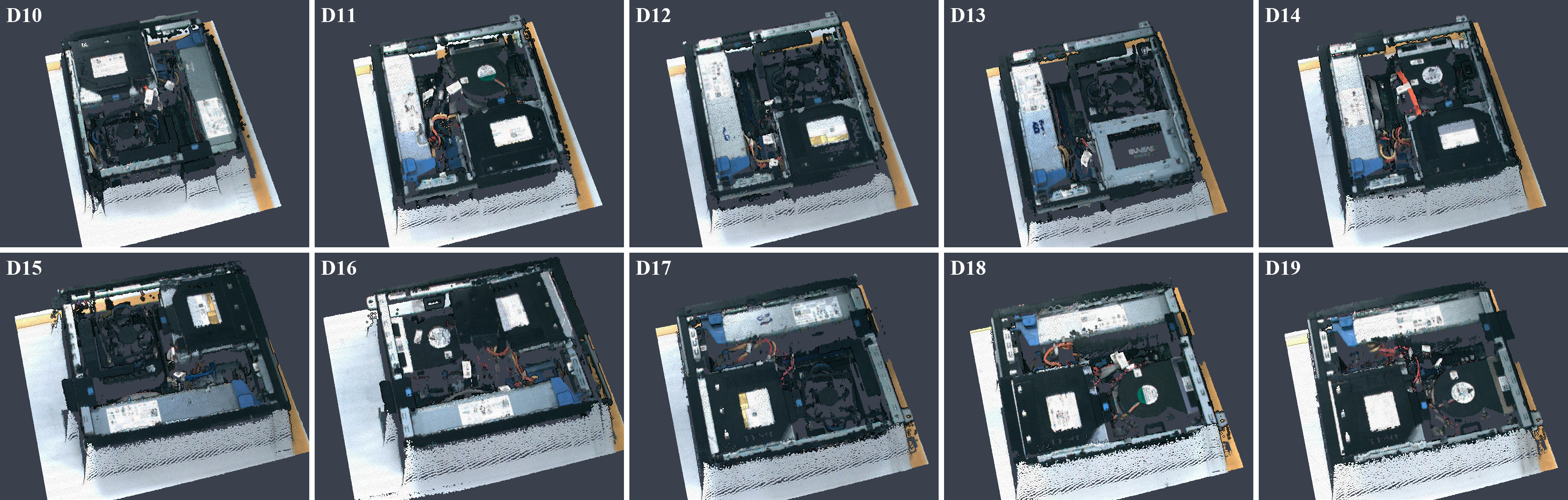}
\caption{\textbf{Reconstructions of ten desktops at the initial observation.} The full colored reconstructions of D10 to D19 are cropped to the annotated desktop region and rendered from oblique viewpoints as in \cref{fig:app_pointcloud}.}
\label{fig:app_recon}
\end{figure*}

We use fringe projection profilometry to reconstruct the visible surfaces of the remaining assembly at every disassembly step. Because the RGB image of the Scanner view is recovered from the same captured images, the image and the reconstruction share the scanner camera and its pixel grid. Each reconstructed point maps to one image pixel and carries that pixel's color, forming a colored point cloud (\cref{fig:app_pointcloud,fig:app_recon}). We use each part's visible-region mask to select its surface points as a part-level point cloud. We also project the whole reconstruction onto the image plane with the calibrated scanner intrinsics to obtain the metric reference depth maps used in \cref{sec:app_depth_qual}.

The released part-level point clouds are filtered by class and reconstruction coverage. The coverage of a part in an observation is the number of reconstructed points under its visible-region mask divided by the number of pixels in that mask. We exclude the chassis, storage drives, case connector parts, and screw holes. A part's point cloud is also excluded from any observation where its coverage is below 30\%. For connectors, screws, and RAM latches, we keep at most three parts per class and desktop. This filtering gives 31,705 part-level point clouds.

\section{QA Construction and Scoring}
\label{sec:app_qa}

\subsection{Task Construction}

Template code generates every question and its reference answer from the source records. Perception answers come from the masks, state records, and disassembly step associated with each image. Understanding answers come from the operation logs and changes between observations. Planning answers come from the disassembly graph, current states, and the four conditions in \cref{sec:suppl_graph}. The QA set has 17 task forms (\cref{tab:app_qa_counts}). Twelve are the primary tasks in \cref{tab:level2}. The other five are additional forms of primary tasks: Observed operation with a region hint, Operations across several steps, State change across several steps, and the constrained and multiple-choice forms of Next feasible action categories. Same step across views has one row for each of three groups of camera pairs. Camera pairs are grouped by the view that comes first in the fixed order Scanner view, OAK-D top view, OAK-D side view, RealSense view. The primary task in \cref{tab:level2} uses the Scanner group.

\begin{table*}[!tp]
\caption{\textbf{The 17 task forms and their counts.} The split columns count the generated QA pairs by desktop split. The Evaluated column counts the test questions of the evaluation sample (\cref{sec:app_vlm_sample}). Bullets identify the twelve primary tasks in \cref{tab:level2}. The final cell in the Primary column gives the number of evaluated questions in the primary tasks.}
\label{tab:app_qa_counts}
\centering
\footnotesize
\setlength{\tabcolsep}{5pt}
\begin{tabular}{@{}llrrrrc@{}}
\toprule
& & \multicolumn{3}{c}{QA pairs by split} & & \\
\cmidrule(lr){3-5}
Task & Images given & Train & Validation & Test & Evaluated & Primary \\
\midrule
\multicolumn{7}{@{}l}{\textbf{Perception}} \\
Visible component boxes & one image & 2,057 & 256 & 491 & 55 & $\bullet$ \\
Component grounding & one image & 21,365 & 2,675 & 5,106 & 47 & $\bullet$ \\
Component state or screw count & one image, part outlined & 25,313 & 3,438 & 5,598 & 54 & $\bullet$ \\
Same step across views, Scanner pairs & two views & 11,082 & 1,428 & 2,784 & 37 & $\bullet$ \\
Same step across views, OAK-D top pairs & two views & 7,394 & 946 & 1,848 & 24 &  \\
Same step across views, OAK-D side pairs & two views & 3,680 & 468 & 920 & 12 &  \\
\midrule
\multicolumn{7}{@{}l}{\textbf{Understanding}} \\
Observed operation & all views, before and after & 1,703 & 217 & 411 & 54 & $\bullet$ \\
Observed operation, region hint & same, with a hint box & 1,703 & 217 & 411 & 54 &  \\
Operations across several steps & all views, 2--3 steps apart & 1,362 & 189 & 374 & 51 &  \\
State change & all views, before and after & 1,983 & 248 & 477 & 74 & $\bullet$ \\
State change across several steps & all views, 2--3 steps apart & 3,835 & 482 & 922 & 74 &  \\
Completed operations & all views, current step & 1,934 & 215 & 492 & 58 & $\bullet$ \\
\midrule
\multicolumn{7}{@{}l}{\textbf{Planning}} \\
Proposed action feasibility & Scanner image, part outlined & 28,533 & 3,461 & 6,494 & 71 & $\bullet$ \\
Next feasible action categories & all views, current step & 1,780 & 225 & 407 & 52 & $\bullet$ \\
Next feasible action categories, constrained & all views, current step & 784 & 141 & 184 & 53 &  \\
Next feasible action categories, multiple choice & all views, current step & 1,024 & 143 & 213 & 23 &  \\
Target removal plan & Scanner image, target outlined & 5,238 & 676 & 1,272 & 53 & $\bullet$ \\
Constrained removal plan & Scanner image, target outlined & 4,930 & 624 & 1,198 & 51 & $\bullet$ \\
Proposed plan validity & Scanner image, parts outlined & 9,375 & 1,243 & 2,294 & 57 & $\bullet$ \\
\midrule
Total & & 135,075 & 17,292 & 31,896 & 954 & 663 \\
\bottomrule
\end{tabular}
\end{table*}

\paragraph{Same step across views.}
In Same step across views, a question shows images from two cameras and asks whether they show the same disassembly step. The task covers all six camera pairs. In a positive question, both images come from the same disassembly step. In a negative question, the second camera's image comes from the nearest disassembly step, at most six steps away, at which a large part differs. The relevant classes are PSU, motherboard, drive cage, optical drive, cover, CPU cooler, expansion card, and RAM module. When no such step exists, neither the positive nor the negative question is generated. Positive and negative questions are equal in number for each camera pair.

\paragraph{State and state change.}
In Component state or screw count, the reference answer is \texttt{cannot\_tell} when a part is in place but cannot be seen in the shown view because it is fully hidden, too small, or outside the view. Such a question concerns either the only part of its class on the desktop or a group whose remaining members are all hidden. In State change across several steps, a no-change question is asked only when the images cannot show a change between the two observations. Every intervening disassembly step must either record another observation without an operation or only loosen captive screws, which stay in place.

\paragraph{Interchangeable parts and plans.}
Some questions do not distinguish between individual parts, such as screws that fasten the same part. These parts form an interchangeable group, and naming any member of the group is correct. Screws are grouped by the part they fasten, connectors by the part they plug into and their cable, and latches by the part they lock. Next feasible action categories uses broader operation categories. All connectors form one category, all RAM latches form another, and all RAM modules form another. Screws are grouped by the part they fasten, and every other part is its own category. Alternative operations on one part count once. For Proposed plan validity, an invalid plan is made by substituting, moving, or swapping one operation in a valid plan.

\subsection{Question Inputs and Review}

A question refers to a part by its class name or by a thin dashed red outline with a letter tag drawn on the image. Interchangeable parts can share a tag. Questions with outlined parts show only the Scanner image. For questions that show all available views, the images of each observation are ordered Scanner, OAK-D top, OAK-D side, and RealSense. When two observations are shown, all images of the earlier observation precede those of the later observation. Each task has between five and ten question wordings.

All models receive the system message: ``You are shown photographs of a desktop PC being disassembled on a bench. Answer exactly in the JSON format asked for.'' The user message contains the view legend, outlined-part list, question, any multiple-choice options, required answer format, and glossary, in that order. The glossary defines relevant classes and states. \Cref{tab:app_qa_examples} gives examples.

\begin{table*}[!tp]
\caption{\textbf{Example questions and reference answers} from the test desktops. S$k$ denotes the $k$-th observation of a desktop, and S01 is the initial observation. Long answers are abbreviated.}
\label{tab:app_qa_examples}
\centering
\footnotesize
\setlength{\tabcolsep}{4pt}
\begin{tabular}{@{}>{\raggedright\arraybackslash}p{0.16\textwidth}>{\raggedright\arraybackslash}p{0.08\textwidth}>{\raggedright\arraybackslash}p{0.37\textwidth}>{\raggedright\arraybackslash}p{0.33\textwidth}@{}}
\toprule
Task & Images & Question & Reference answer \\
\midrule
Visible component boxes & D04, S38 & Name the parts visible in this photo of the desktop PC and localise them. & 14 parts with boxes, e.g.\ \texttt{\{"class": "motherboard", "bbox": [691, 390, 497, 608]\}} \\
Component grounding & D31, S02 & Localise the cable clip in this image. & \texttt{\{"bbox": [993, 647, 47, 38]\}} \\
Component state & D65, S34 & What is the state of the card latch outlined in red? & \texttt{\{"state": "open"\}} \\
Component state & D10, S08 & What is the state of the SSD storage drive outlined in red? & \texttt{\{"state": "cannot\_tell"\}} \\
Screw count & D46, S15 & How many motherboard screws are still in place? & \texttt{\{"count": 10\}} \\
Same step across views & D10, S28 and S34 & Do these two views show the same disassembly state of the same machine? & \texttt{\{"same\_moment": false\}} \\
\midrule
Observed operation & D24, S01 and S02 & Compare the machine before and after the action: which action was carried out? Also say where the machine changed: give that region as a box on the top-down scanner image. & \texttt{\{"verb": "release", "target\_class": "cable\_clip", "tool": "none", "region": [961, 552, 54, 44]\}} \\
Observed operation, region hint & D15, S19 and S20 & Look inside the yellow dashed box on the scanner images: something there changed between before and after. Which action was carried out? & \texttt{\{"verb": "remove", "target\_class": "cpu", "tool": "none"\}} \\
Operations across several steps & D30, S13 and S15 & Between the first moment and the second, the operator carried out several actions. Which ones? & Disconnect one of the connectors, remove the PSU \\
State change & D48, S08 and S09 & Do the pictures before and after show the same machine state, or did something change? & \texttt{\{"changed": true\}} \\
State change across several steps & D04, S03 and S06 & Between the earlier and the later pictures, did anything about the machine change? If so, what? & \texttt{\{"changed": false\}} \\
Completed operations & D65, S04 & Which disassembly actions have already been carried out on this machine? & Done: disconnect the case fan connector, remove the case fan. Progress: 0--25\% \\
\midrule
Proposed action feasibility & D46, S13 & Can you remove the CPU outlined in red right now? If not, what has to be dealt with first? & \texttt{\{"feasible": false, "blockers": ["CPU socket lever"]\}} \\
Next feasible action categories & D20, S37 & Looking at the machine in these pictures, which actions could be carried out right now? List all of them. & Remove the cover, remove the PSU latch, unscrew the motherboard screws \\
Next feasible action categories, constrained & D31, S43 & Name every action that is possible on this machine as it stands, keeping to this rule: The optical drive cage has to come out before the cover is taken out. & Remove the optical drive cage \\
Next feasible action categories, multiple choice & D04, S31 & Only one of these actions can be done on the machine as it stands. Which one? A.~take out the cover, B.~take out the motherboard, C.~take out the PSU & \texttt{\{"choice": "A"\}} \\
Target removal plan & D10, S11 & Looking at the machine as it stands, give the ordered list of disassembly actions that have to happen for the fan CPU cooler outlined in red to come out. & 1.~unscrew the CPU cooler screws, 2.~disconnect the fan connector, 3.~remove the CPU cooler \\
Constrained removal plan & D65, S28 & Plan the steps to take the GPU expansion card outlined in red out of this machine, starting from the machine as it is now, under this rule: The optical drive cage has to come out before the GPU expansion card is taken out. List the actions in order. & 1.~open the card latch, 2.~remove the optical drive cage, 3.~remove the GPU expansion card \\
Proposed plan validity & D24, S32 & Verify the following sequence. Say whether it can be carried out and, if not, the index of the first impossible action. 1.~remove part A, 2.~displace part B, 3.~remove part B & \texttt{\{"valid": false, "first\_error\_step": 2\}} \\
\bottomrule
\end{tabular}
\end{table*}

Reasoning traces state the observation, applicable rule, and derivation of the reference answer. They are generated from the same records as the answers. They are available for training.

After the manual review and revision described in the main paper, three automatic checks run on the whole QA set. An independently written second implementation recomputes every reference answer, and the two implementations agree. The checks also test conditions that must always hold, including the feasibility of every operation in the operation logs and successful replay of every reference plan. No tested condition is violated. The scorer also grades the reference answers themselves, and every one receives full credit.

\subsection{Scoring}
\label{sec:app_vlm_scoring}

\Cref{tab:tasks} lists the metric for each task. All metrics use strict parsing, which requires the entire reply to be a JSON object in the requested format. A reply that cannot be parsed scores zero and stays in the denominator. To measure the effect of strict parsing on scores (\cref{sec:app_vlm_parsing}), we also score replies with a lenient rule that extracts the last JSON object in the reply. If no JSON object is present, the rule accepts yes or no, an option letter, an integer, or a state word where the task permits it.

\begin{table*}[!tp]
\caption{\textbf{Scoring rules of the QA tasks.} All scores lie between 0 and 1 and are reported on a 0--100 scale.}
\label{tab:tasks}
\centering\small
\setlength{\tabcolsep}{4pt}
\begin{tabular}{@{}p{.24\textwidth}p{.72\textwidth}@{}}
\toprule
Task & Scoring rule \\
\midrule
Visible component boxes & Predicted boxes are matched one-to-one to annotated boxes of the same class at IoU $\geq 0.50$. For $M$ matches, $P$ predicted boxes, and $G$ annotated boxes, the box-set F1 is $2M/(P+G)$. \\
Component grounding & For one target, the score is 1 if the box has IoU $\geq 0.50$ with the annotated box and 0 otherwise. For all parts of a class, the score is the box-set F1. \\
Same step across views & Accuracy of the same-step decision. \\
Component state or screw count & A state must match exactly, including \texttt{cannot\_tell}. A correct count scores 1, an error of one scores 0.8, and a larger error scores 0. \\
Observed operation & The operation and the target class must both be correct. The region-hint form uses the same rule. The several-step form scores the set F1 over the groups of operations carried out between the two observations. \\
State change & Accuracy in deciding whether a visible state change occurred. The several-step form uses the same rule. \\
Completed operations & Set F1 over the groups of operations completed up to the current observation. \\
Proposed action feasibility & Accuracy of the feasibility decision. \\
Next feasible action categories & Each answered operation is mapped to its operation category. Let $H$ be the number of distinct feasible operation categories named, $W$ the number of answered operations that are not feasible, and $G$ the number of feasible operation categories. Precision is $H/(H+W)$, recall is $H/G$, and the score is their F1. The constrained form adds a rule to the question. The multiple-choice form scores the choice of the single feasible option. \\
Target removal plan & The plan is replayed from the current state. It succeeds if the target is removed, no step violates a prerequisite, and all required operations are included. Additional feasible steps are not penalized. \\
Constrained removal plan & Same as the target removal plan, with an additional rule in the question that the plan must respect. \\
Proposed plan validity & Accuracy of the decision whether the proposed plan can be carried out. \\
\bottomrule
\end{tabular}
\end{table*}

\paragraph{Boxes and counts.}
Boxes are requested in pixel coordinates of the image. Some models answer in an internally resized frame at 0.8 of the image size. Multiplying these coordinates by 1.25 maps them back to image pixels. For every model, each answer to Visible component boxes and Component grounding is scored twice, once as given and once with all box coordinates multiplied by 1.25. The higher score for the whole answer is kept. A count that is not an integer scores 0.

\paragraph{Operation sets.}
Set F1 is 1 when both sets are empty and 0 when only one is empty. For Next feasible action categories and Operations across several steps, the operations in a reply are mapped to operation categories. Repeating a category that is already counted has no penalty. Operations on blocked, already handled, or absent parts, and operations that match no part, count against precision. When a reply names only a class, each such operation is matched to a different part of that class in the reference answer. An invalid multiple-choice letter scores 0. For Completed operations, wording that matches no category is discarded without penalty.

\paragraph{Operation recognition and plan replay.}
For Observed operation, when a disassembly step contains several operations, the best-matching operation determines the score. Removal plans are evaluated against the operations still required for the target, not against the order in the operation log. Replay skips unmatched steps, repeated steps, and steps on already handled parts without penalty. An operation on an interchangeable group acts on every member. A plan succeeds when it reaches the goal, completes all required operations, and violates no prerequisite or stated rule.

Some requested answer fields are not scored: the tool and region in Observed operation, the blockers in Proposed action feasibility, the started operations and the progress range in Completed operations, the first invalid step in Proposed plan validity, and the reasons in Next feasible action categories.

\section{Component Detection and Segmentation}
\label{sec:app_perception_tasks}

\begin{table*}[!tp]
\caption{\textbf{Detection and segmentation systems.} Checkpoints are selected by mask AP on the validation split, except for the fine-tuned SAM~3, which uses its final checkpoint.}
\label{tab:app_vision_impl}
\centering
\footnotesize
\setlength{\tabcolsep}{4pt}
\begin{tabular}{@{}llrll@{}}
\toprule
System & Trained parameters & Input size & Schedule & Selected checkpoint \\
\midrule
YOLO26m-seg & All & 1600 & 100 epochs & Epoch 20 \\
Mask R-CNN R50-FPN v2 & All & 1600 & 24 epochs, batch size 2 & Epoch 12 \\
DINOv3 + Mask R-CNN & Feature pyramid and heads & 1600 & 24 epochs & Epoch 24 \\
RF-DETR-Seg & All & 1608 & 100 epochs, effective batch size 16 & Best of every tenth epoch \\
RF-DETR + frozen SAM 2.1 & None & -- & -- & -- \\
RF-DETR + frozen SAM 3 & None & -- & -- & -- \\
SAM 3, zero-shot & None & -- & -- & -- \\
SAM 3, fine-tuned & All (840.5~M) & 1008 & 20 epochs, batch size 2 & Final epoch \\
\bottomrule
\end{tabular}
\end{table*}

\begin{table*}[!tp]
\caption{\textbf{Results by instance size and at the selected operating point.} Average precision for small, medium, and large instances (AP$_S$, AP$_M$, AP$_L$) uses COCO area bins. Macro-F1 averages F1 over the 21 supported classes. Micro-F1 pools detections and includes a 95\% bootstrap confidence interval (CI). At the validation-selected threshold and intersection over union 0.50, P and R denote micro-precision and micro-recall. TP, FP, and FN count true positives, false positives, and false negatives.}
\label{tab:vision_size_counts}
\centering\footnotesize
\setlength{\tabcolsep}{1.8pt}
\begin{tabular*}{\textwidth}{@{}l@{\extracolsep{\fill}}rrrrrrrrrrr@{}}
\toprule
& \multicolumn{5}{c}{Scores $\uparrow$} & \multicolumn{3}{c}{Operating point} & \multicolumn{3}{c}{Instance counts} \\
\cmidrule(lr){2-6}\cmidrule(lr){7-9}\cmidrule(l){10-12}
Method & AP$_S$ & AP$_M$ & AP$_L$ & Macro-F1 & Micro-F1 [95\% CI] & Threshold & P & R & TP & FP & FN \\
\midrule
\multicolumn{12}{@{}l}{\textit{Bounding boxes}} \\
YOLO26m-seg & 14.6 & 27.0 & 53.3 & 64.2 & 76.0 [71.0, 79.9] & 0.141 & 88.0 & 66.8 & 8,642 & 1,182 & 4,288 \\
Mask R-CNN R50-FPN v2 & 13.4 & 27.4 & 63.3 & 66.1 & 76.4 [71.9, 80.4] & 0.384 & 84.9 & 69.5 & 8,989 & 1,600 & 3,941 \\
DINOv3 + Mask R-CNN & 12.6 & 16.5 & 52.5 & 63.8 & 73.3 [68.2, 77.7] & 0.578 & 85.4 & 64.3 & 8,309 & 1,418 & 4,621 \\
RF-DETR-Seg & 14.9 & 25.9 & 69.9 & 71.0 & 77.2 [72.7, 81.1] & 0.459 & 88.5 & 68.4 & 8,849 & 1,152 & 4,081 \\
SAM 3, zero-shot & 0.4 & 1.7 & 6.0 & 6.0 & 7.0 [5.8, 8.3] & 0.136 & 4.5 & 16.6 & 2,147 & 45,881 & 10,783 \\
SAM 3, fine-tuned & 16.1 & 33.3 & 72.4 & 75.2 & 76.2 [72.0, 80.0] & 0.508 & 78.8 & 73.9 & 9,550 & 2,570 & 3,380 \\
\midrule
\multicolumn{12}{@{}l}{\textit{Visible-region masks}} \\
YOLO26m-seg & 14.2 & 25.1 & 51.4 & 64.0 & 75.8 [70.4, 80.1] & 0.141 & 87.8 & 66.7 & 8,627 & 1,202 & 4,303 \\
Mask R-CNN R50-FPN v2 & 12.6 & 24.5 & 54.8 & 64.5 & 75.0 [70.2, 79.2] & 0.408 & 83.5 & 68.0 & 8,798 & 1,738 & 4,132 \\
DINOv3 + Mask R-CNN & 11.7 & 14.5 & 46.7 & 62.8 & 72.4 [66.8, 77.3] & 0.578 & 84.2 & 63.5 & 8,206 & 1,539 & 4,724 \\
RF-DETR-Seg & 11.7 & 23.1 & 62.4 & 70.0 & 75.2 [70.8, 79.0] & 0.459 & 86.2 & 66.7 & 8,625 & 1,378 & 4,305 \\
RF-DETR + SAM 2.1 & 8.3 & 17.6 & 44.0 & 58.8 & 62.2 [58.7, 65.7] & 0.459 & 71.3 & 55.1 & 7,130 & 2,873 & 5,800 \\
RF-DETR + SAM 3 & 9.3 & 18.1 & 44.3 & 59.8 & 64.6 [61.2, 68.1] & 0.459 & 74.1 & 57.3 & 7,411 & 2,592 & 5,519 \\
SAM 3, zero-shot & 0.3 & 1.7 & 4.8 & 5.9 & 6.7 [5.4, 8.0] & 0.136 & 4.3 & 15.8 & 2,047 & 46,021 & 10,883 \\
SAM 3, fine-tuned & 14.1 & 31.6 & 66.3 & 73.9 & 75.6 [71.0, 79.6] & 0.508 & 78.0 & 73.3 & 9,474 & 2,673 & 3,456 \\
\bottomrule
\end{tabular*}
\end{table*}

\subsection{Evaluation Protocol}

The benchmark uses 492 Scanner images from the test desktops and a 24-class vocabulary. The test images contain 12,930 instances with visible-region masks. Cooler bracket, case connector part, and screw hole have no test instances, and the other 21 classes are the supported classes. All predictions are evaluated in the $1600\times1600$ Scanner image frame.

A fully hidden part has no visible pixels. Its annotation keeps the full-extent box and an empty visible-region mask and is treated as a COCO crowd region. A box detection of the same class that falls within this box is ignored in AP and F1. The empty mask defines no ignore region, so a mask predicted for a fully hidden part counts as a false positive.

COCO AP averages IoU thresholds from 0.50 to 0.95 in steps of 0.05. Micro-F1 matches same-class predictions and annotations at IoU 0.50 after confidence filtering. Each system's box and mask thresholds are selected separately on validation predictions. Macro-F1 averages over the supported classes.

For pixel overlap, masks of the same class are merged within each image. Class intersections and unions are summed across the test set. Class mIoU averages over supported classes. Pixel IoU divides total intersection by total union over all 24 classes, so predictions of unsupported classes count against it.

Confidence intervals use 1,000 fixed-seed bootstrap resamples of the 12 test desktops with replacement. The confidence threshold stays fixed. We report percentile intervals.

\begin{table*}[!tp]
\caption{\textbf{Box and mask AP by occlusion ratio.} Each cell gives average precision (AP) across intersection over union thresholds 0.50--0.95 and at 0.50 (AP$_{50}$), on a 0--100 scale. The four bins contain 9,655, 2,084, 583, and 608 annotated instances, respectively.}
\label{tab:suppl_occlusion}
\centering\footnotesize
\setlength{\tabcolsep}{2pt}
\begin{tabular*}{\textwidth}{@{}l@{\extracolsep{\fill}}rrrrrrrr@{}}
\toprule
& \multicolumn{4}{c}{Detection $\uparrow$} & \multicolumn{4}{c}{Instance segmentation $\uparrow$} \\
\cmidrule(lr){2-5}\cmidrule(l){6-9}
Occlusion ratio & $[0,.01)$ & $[.01,.30)$ & $[.30,.60)$ & $[.60,1]$ & $[0,.01)$ & $[.01,.30)$ & $[.30,.60)$ & $[.60,1]$ \\
\midrule
YOLO26m-seg & 48.5 / 71.2 & 32.9 / 50.5 & 10.5 / 20.2 & 8.7 / 16.7 & 45.1 / 66.3 & 32.1 / 45.1 & 10.3 / 19.0 & 6.7 / 14.8 \\
Mask R-CNN R50-FPN v2 & 48.9 / 72.0 & 41.5 / 59.1 & 26.2 / 40.1 & 15.1 / 24.1 & 45.6 / 66.3 & 37.0 / 52.6 & 15.7 / 29.2 & 5.9 / 16.8 \\
DINOv3 + Mask R-CNN & 40.5 / 70.8 & 37.1 / 56.6 & 14.5 / 23.4 & 12.1 / 19.7 & 37.2 / 65.5 & 33.3 / 52.1 & 8.1 / 18.5 & 3.9 / 14.4 \\
\midrule
RF-DETR-Seg & \multirow[c]{3}{*}{55.6 / 74.1} & \multirow[c]{3}{*}{47.7 / 59.5} & \multirow[c]{3}{*}{26.5 / 39.8} & \multirow[c]{3}{*}{14.2 / 17.8} & 49.0 / 72.9 & 42.8 / 56.5 & 17.7 / 34.1 & 11.6 / 18.6 \\
RF-DETR + SAM 2.1 &  &  &  &  & 34.3 / 53.6 & 30.4 / 48.8 & 6.5 / 25.0 & 1.4 / 2.0 \\
RF-DETR + SAM 3 &  &  &  &  & 36.3 / 57.4 & 30.7 / 49.5 & 6.9 / 25.5 & 1.4 / 2.0 \\
\midrule
SAM 3, zero-shot & 0.3 / 0.7 & 7.6 / 8.2 & 0.4 / 1.2 & 0.0 / 0.1 & 0.3 / 0.7 & 5.8 / 8.2 & 0.5 / 1.2 & 0.0 / 0.1 \\
SAM 3, fine-tuned & 64.1 / 81.6 & 47.9 / 60.7 & 21.2 / 32.5 & 16.5 / 21.2 & 56.8 / 80.6 & 42.5 / 59.3 & 17.7 / 26.7 & 12.1 / 20.4 \\
\bottomrule
\end{tabular*}
\end{table*}

\begin{table*}[!tp]
\caption{\textbf{Visible-region mask AP by component class.} Instance counts give test support. RF + SAM denotes a frozen SAM refiner prompted with RF-DETR-Seg boxes. ZS and FT denote zero-shot and fine-tuned text-prompted SAM~3. A dash indicates no test instances.}
\label{tab:suppl_class}
\centering\footnotesize
\setlength{\tabcolsep}{3pt}
\begin{tabular}{@{}lrrrrrrrrr@{}}
\toprule
& & \multicolumn{8}{c}{Mask AP $\uparrow$} \\
\cmidrule(l){3-10}
Component class & Instances & YOLO & Mask R-CNN & DINOv3 & RF-DETR & RF + SAM 2.1 & RF + SAM 3 & SAM 3 ZS & SAM 3 FT \\
\midrule
chassis & 492 & 44.7 & 52.3 & 39.7 & 76.0 & 1.1 & 1.1 & 0.0 & 70.7 \\
drive cage & 304 & 47.2 & 63.7 & 48.6 & 70.9 & 56.9 & 54.7 & 0.0 & 65.9 \\
cover & 433 & 59.1 & 61.5 & 57.3 & 63.9 & 61.9 & 59.4 & 0.0 & 72.9 \\
cooler bracket & 0 & -- & -- & -- & -- & -- & -- & -- & -- \\
motherboard & 478 & 72.9 & 74.5 & 69.5 & 88.1 & 53.7 & 57.3 & 60.8 & 86.2 \\
cpu & 181 & 86.9 & 84.7 & 81.3 & 86.9 & 68.0 & 63.6 & 3.0 & 83.3 \\
cpu cooler & 135 & 58.6 & 73.3 & 65.2 & 79.2 & 70.7 & 71.5 & 11.2 & 78.3 \\
ram module & 416 & 20.2 & 21.7 & 14.0 & 19.9 & 1.6 & 1.4 & 0.0 & 17.0 \\
psu & 279 & 68.7 & 76.8 & 70.4 & 88.0 & 60.0 & 69.8 & 0.0 & 85.2 \\
storage drive & 101 & 30.1 & 26.9 & 29.2 & 37.5 & 37.0 & 38.3 & 0.0 & 48.3 \\
optical drive & 32 & 85.9 & 88.1 & 81.4 & 97.5 & 87.5 & 81.9 & 0.0 & 97.0 \\
expansion card & 180 & 8.9 & 17.6 & 14.0 & 21.9 & 11.2 & 11.6 & 0.0 & 38.9 \\
case fan & 2 & 1.3 & 5.1 & 0.7 & 6.6 & 10.0 & 9.0 & 1.6 & 70.1 \\
case connector part & 0 & -- & -- & -- & -- & -- & -- & -- & -- \\
screw & 3,065 & 35.6 & 31.6 & 30.6 & 29.7 & 8.9 & 13.2 & 2.3 & 33.8 \\
ram latch & 3,148 & 49.9 & 45.7 & 45.5 & 41.0 & 42.2 & 43.4 & 0.0 & 51.8 \\
cpu socket lever & 380 & 57.5 & 50.2 & 37.0 & 56.8 & 1.7 & 1.8 & 0.0 & 55.9 \\
psu latch & 138 & 88.3 & 85.6 & 59.5 & 89.8 & 87.7 & 88.0 & 0.1 & 93.9 \\
drive latch & 49 & 65.2 & 67.3 & 45.8 & 59.6 & 39.5 & 40.7 & 0.0 & 68.4 \\
card latch & 43 & 0.0 & 0.0 & 2.0 & 2.1 & 1.9 & 1.9 & 0.7 & 1.9 \\
cooler latch & 1 & 0.0 & 0.0 & 0.0 & 10.0 & 10.0 & 10.0 & 0.0 & 0.0 \\
cable clip & 113 & 12.5 & 2.2 & 1.3 & 4.5 & 5.2 & 5.5 & 0.0 & 11.3 \\
connector & 2,960 & 17.6 & 14.2 & 9.0 & 12.2 & 9.2 & 10.4 & 0.1 & 17.4 \\
screw hole & 0 & -- & -- & -- & -- & -- & -- & -- & -- \\
\bottomrule
\end{tabular}
\end{table*}

\subsection{Training and Model Inputs}

Trained systems use Scanner images and visible-region masks from the training desktops, with the settings in \cref{tab:app_vision_impl}. Each has one training run on NVIDIA A100 GPUs. YOLO26m-seg, Mask R-CNN R50-FPN v2, and RF-DETR-Seg are trained end to end.

\paragraph{YOLO and Mask R-CNN.}
YOLO26m-seg uses batch size 1, mixed precision, and seed 0. Mask R-CNN uses batch size 2 and mixed precision for 24,624 iterations.

\paragraph{DINOv3 features.}
The DINOv3 system freezes its DINOv3-pretrained ConvNeXt-Tiny backbone and trains the feature pyramid and Mask R-CNN heads. Training uses SGD with learning rate 0.02, momentum 0.9, weight decay $10^{-4}$, batch size 2, seed 0, and no mixed precision.

\paragraph{RF-DETR-Seg.}
The Small variant starts from its released segmentation checkpoint. The learning rate is $10^{-4}$, with $1.5\times10^{-4}$ for the encoder, weight decay $10^{-4}$, and a learning-rate drop at epoch 80. Batch size is one image with 16 accumulation steps. Training uses gradient checkpointing, without warm-up or an exponential moving average. No augmentation is used beyond a square resize to 1608. Multi-scale training is disabled.

\paragraph{SAM systems.}
Zero-shot and fine-tuned SAM~3 use the same 24 class names as text prompts. Fine-tuning updates all parameters. Learning rates are $8\times10^{-5}$ for the transformer, $2.5\times10^{-5}$ for the vision backbone, and $5\times10^{-6}$ for the language backbone, with weight decay 0.1. Training uses batch size 1 on each of 2 GPUs, layer-wise vision-backbone learning-rate decay 0.9, gradient clipping at 0.1, bfloat16, seed 123, and input size 1008.

The frozen refiners retain RF-DETR-Seg boxes, class labels, and scores and replace only the masks. They prompt the released SAM~2.1 Hiera Base+ or SAM~3 checkpoint with the predicted boxes.

\begin{figure*}[!tp]
\centering
\includegraphics[width=\textwidth]{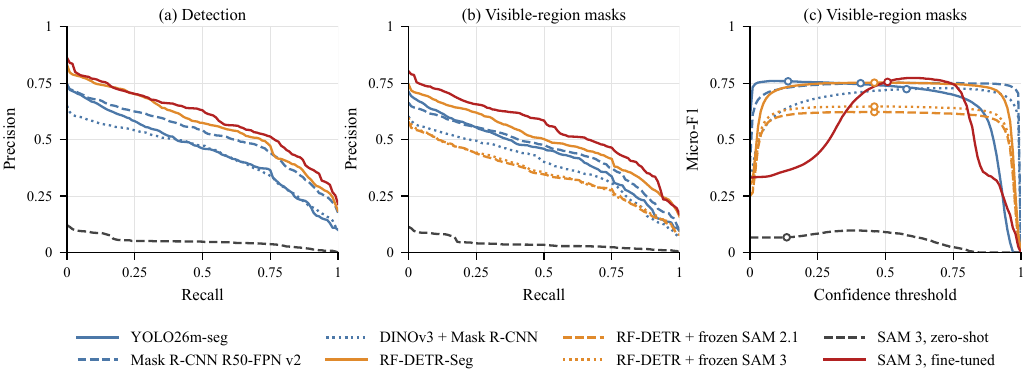}
\caption{\textbf{Evaluation curves of all eight systems.} (a) Detection and (b) visible-region mask precision--recall curves, averaged over IoU thresholds from 0.50 to 0.95. The frozen refiners keep the RF-DETR-Seg boxes, so their detection curves coincide with that of RF-DETR-Seg. (c) Micro-F1 of the visible-region masks at IoU 0.50 as the confidence threshold varies. Circles mark the thresholds selected on the validation split. All curves use the same 492 test images.}
\label{fig:vision_curves}
\end{figure*}

\subsection{Additional Results}

\Cref{tab:vision_size_counts} reports AP by instance size together with each system's operating point. \Cref{tab:suppl_occlusion} reports AP in four bins of the occlusion ratio. Its first and last bins give AP$_{\mathrm{vis}}$ and AP$_{\mathrm{occ}}$ in \cref{tab:level1}.

Performance differs even between frequently observed classes of small parts (\cref{tab:suppl_class}). Fine-tuned SAM~3 obtains 33.8 mask AP for screws and 17.4 for connectors. The frozen refiners perform poorly on chassis and CPU socket lever masks.

\Cref{fig:vision_curves} shows that fine-tuned SAM~3 has the highest precision among the eight systems over most of the recall range, for boxes and for masks. Both frozen refiners have lower mask precision than RF-DETR-Seg at every recall. Zero-shot SAM~3 has lower precision than every trained system at every recall.

\section{Vision-Language Reasoning}
\label{sec:app_vlm}

\begin{table*}[!tp]
\caption{\textbf{Results on all 17 task forms for the 954 test questions.} Scores use strict parsing and a 0--100 scale. ZS and FT denote the zero-shot and fine-tuned models. $^\dagger$The prior-incorrect row is the subset of Component state or screw count whose reference answer is not the most frequent state of the part's class.}
\label{tab:suppl_tasks}
\centering\footnotesize
\setlength{\tabcolsep}{4pt}
\begin{tabular}{@{}lrrrrrrrrr@{}}
\toprule
& & \multicolumn{4}{c}{With images $\uparrow$} & \multicolumn{4}{c}{No image $\uparrow$} \\
\cmidrule(lr){3-6}\cmidrule(l){7-10}
& & \multicolumn{2}{c}{Qwen} & \multicolumn{2}{c}{InternVL} & \multicolumn{2}{c}{Qwen} & \multicolumn{2}{c}{InternVL} \\
Task & Questions & ZS & FT & ZS & FT & ZS & FT & ZS & FT \\
\midrule
\multicolumn{10}{@{}l}{\textbf{Perception}} \\
Visible component boxes & 55 & 2.5 & 18.9 & 0.6 & 17.9 & --- & --- & --- & --- \\
Component grounding & 47 & 2.1 & 51.1 & 4.3 & 51.1 & --- & --- & --- & --- \\
Component state or screw count & 54 & 35.2 & 83.3 & 38.5 & 83.3 & --- & --- & --- & --- \\
State or count, prior incorrect$^\dagger$ & 29 & 24.1 & 75.9 & 16.6 & 79.3 & --- & --- & --- & --- \\
Same step across views, Scanner pairs & 37 & 43.2 & 86.5 & 48.6 & 83.8 & --- & --- & --- & --- \\
Same step across views, OAK-D top pairs & 24 & 58.3 & 79.2 & 58.3 & 66.7 & --- & --- & --- & --- \\
Same step across views, OAK-D side pairs & 12 & 41.7 & 91.7 & 50.0 & 83.3 & --- & --- & --- & --- \\
\addlinespace[2pt]
\multicolumn{10}{@{}l}{\textbf{Understanding}} \\
Observed operation & 54 & 3.7 & 74.1 & 3.7 & 74.1 & --- & --- & --- & --- \\
Observed operation, region hint & 54 & 3.7 & 79.6 & 3.7 & 75.9 & --- & --- & --- & --- \\
Operations across several steps & 51 & 10.4 & 75.1 & 5.5 & 71.8 & --- & --- & --- & --- \\
State change & 74 & 56.8 & 95.9 & 56.8 & 89.2 & --- & --- & --- & --- \\
State change across several steps & 74 & 60.8 & 94.6 & 56.8 & 89.2 & --- & --- & --- & --- \\
Completed operations & 58 & 5.2 & 75.1 & 5.2 & 70.4 & --- & --- & --- & --- \\
\addlinespace[2pt]
\multicolumn{10}{@{}l}{\textbf{Planning}} \\
Proposed action feasibility & 71 & 46.5 & 73.2 & 54.9 & 63.4 & 40.8 & 45.1 & 56.3 & 50.7 \\
Next feasible action categories & 52 & 3.3 & 71.2 & 2.3 & 65.3 & --- & --- & --- & --- \\
Next feasible action categories, constrained & 53 & 8.6 & 73.2 & 5.7 & 70.0 & 3.6 & 45.4 & 7.0 & 43.6 \\
Next feasible action categories, multiple choice & 23 & 73.9 & 87.0 & 0.0 & 82.6 & 60.9 & 87.0 & 65.2 & 82.6 \\
Target removal plan & 53 & 0.0 & 98.1 & 0.0 & 100.0 & 0.0 & 67.9 & 0.0 & 67.9 \\
Constrained removal plan & 51 & 0.0 & 49.0 & 0.0 & 47.1 & 0.0 & 27.5 & 0.0 & 27.5 \\
Proposed plan validity & 57 & 54.4 & 87.7 & 33.3 & 86.0 & 57.9 & 63.2 & 57.9 & 78.9 \\
\bottomrule
\end{tabular}
\end{table*}

\begin{table*}[!tp]
\caption{\textbf{95\% confidence intervals for the scores with images on the test desktops.} Intervals use 1,000 bootstrap resamples of the test desktops with replacement. Counts give questions and contributing desktops. $^\dagger$The prior-incorrect row is a subset of Component state or screw count.}
\label{tab:suppl_intervals}
\centering\footnotesize
\setlength{\tabcolsep}{5pt}
\begin{tabular}{@{}lrrrrrr@{}}
\toprule
& & & \multicolumn{2}{c}{Qwen3-VL-8B} & \multicolumn{2}{c}{InternVL3.5-8B} \\
Task & Questions & Desktops & Zero-shot & Fine-tuned & Zero-shot & Fine-tuned \\
\midrule
\multicolumn{7}{@{}l}{\textbf{Perception}} \\
Visible component boxes & 55 & 12 & 1.2--3.7 & 15.0--23.8 & 0.1--1.1 & 12.7--24.5 \\
Component grounding & 47 & 12 & 0.0--6.7 & 37.5--67.4 & 0.0--10.9 & 37.0--66.7 \\
Component state or screw count & 54 & 12 & 18.4--50.0 & 73.1--92.9 & 27.2--48.4 & 71.2--94.6 \\
State or count, prior incorrect$^\dagger$ & 29 & 12 & 6.9--41.9 & 58.6--89.7 & 3.2--33.1 & 59.3--96.4 \\
Same step across views, Scanner pairs & 37 & 12 & 30.6--55.6 & 73.0--100.0 & 35.0--63.9 & 71.8--94.4 \\
Same step across views, OAK-D top pairs & 24 & 12 & 50.0--70.8 & 66.7--91.7 & 50.0--70.8 & 54.2--79.2 \\
Same step across views, OAK-D side pairs & 12 & 12 & 16.7--66.7 & 75.0--100.0 & 25.0--83.3 & 58.3--100.0 \\
\addlinespace[2pt]
\multicolumn{7}{@{}l}{\textbf{Understanding}} \\
Observed operation & 54 & 12 & 0.0--9.3 & 63.0--83.6 & 0.0--9.3 & 63.0--84.9 \\
Observed operation, region hint & 54 & 12 & 0.0--9.3 & 69.5--88.7 & 0.0--8.8 & 64.0--87.8 \\
Operations across several steps & 51 & 12 & 4.6--18.1 & 66.8--83.5 & 1.1--11.5 & 62.2--82.4 \\
State change & 74 & 12 & 52.7--61.3 & 87.7--100.0 & 46.7--68.0 & 79.7--97.3 \\
State change across several steps & 74 & 12 & 55.4--66.7 & 86.3--100.0 & 44.0--71.2 & 80.6--97.3 \\
Completed operations & 58 & 12 & 0.0--10.3 & 64.3--84.5 & 0.0--10.3 & 61.4--79.0 \\
\addlinespace[2pt]
\multicolumn{7}{@{}l}{\textbf{Planning}} \\
Proposed action feasibility & 71 & 12 & 39.4--52.2 & 60.2--88.1 & 31.8--74.0 & 48.2--82.4 \\
Next feasible action categories & 52 & 12 & 0.9--6.3 & 65.0--77.6 & 1.0--3.8 & 59.9--71.3 \\
Next feasible action categories, constrained & 53 & 12 & 4.5--12.5 & 64.2--83.5 & 2.9--8.5 & 61.0--79.2 \\
Next feasible action categories, multiple choice & 23 & 9 & 57.7--90.0 & 66.7--100.0 & 0.0--0.0 & 63.6--96.2 \\
Target removal plan & 53 & 12 & 0.0--0.0 & 94.3--100.0 & 0.0--0.0 & 100.0--100.0 \\
Constrained removal plan & 51 & 12 & 0.0--0.0 & 34.6--64.2 & 0.0--0.0 & 31.4--61.8 \\
Proposed plan validity & 57 & 12 & 48.2--62.1 & 80.0--94.8 & 24.6--43.9 & 75.0--94.7 \\
\bottomrule
\end{tabular}
\end{table*}

\begin{table*}[!tp]
\caption{\textbf{Closed-model results on the 954 test questions.} Scores use the same strict task metrics as the open-model results. $n$ is the number of questions. The prior-incorrect row is a subset of Component state or screw count.}
\label{tab:suppl_closed954}
\centering\footnotesize
\begin{tabular}{@{}lrrrr@{}}
\toprule
& & \multicolumn{2}{c}{With images} & \multicolumn{1}{c}{No image} \\
\cmidrule(lr){3-4}\cmidrule(l){5-5}
Task & $n$ & GPT-6 Astra & Opus 5.5 & GPT-6 Astra \\
\midrule
\multicolumn{5}{@{}l}{\textbf{Perception}} \\
Visible component boxes & 55 & 9.8 & 8.6 & --- \\
Component grounding & 47 & 52.1 & 34.9 & --- \\
Component state or screw count & 54 & 71.9 & 66.3 & --- \\
State or count, prior incorrect & 29 & 72.4 & 51.7 & --- \\
Same step across views, Scanner pairs & 37 & 78.4 & 70.3 & --- \\
Same step across views, OAK-D top pairs & 24 & 87.5 & 58.3 & --- \\
Same step across views, OAK-D side pairs & 12 & 83.3 & 66.7 & --- \\
\addlinespace[2pt]
\multicolumn{5}{@{}l}{\textbf{Understanding}} \\
Observed operation & 54 & 24.1 & 25.9 & --- \\
Observed operation, region hint & 54 & 24.1 & 22.2 & --- \\
Operations across several steps & 51 & 23.9 & 23.9 & --- \\
State change & 74 & 59.5 & 77.0 & --- \\
State change across several steps & 74 & 66.2 & 75.7 & --- \\
Completed operations & 58 & 28.3 & 20.8 & --- \\
\addlinespace[2pt]
\multicolumn{5}{@{}l}{\textbf{Planning}} \\
Proposed action feasibility & 71 & 70.4 & 66.2 & 53.5 \\
Next feasible action categories & 52 & 8.4 & 10.5 & --- \\
Next feasible action categories, constrained & 53 & 19.6 & 9.9 & 0.0 \\
Next feasible action categories, multiple choice & 23 & 21.7 & 21.7 & 43.5 \\
Target removal plan & 53 & 0.0 & 1.9 & 0.0 \\
Constrained removal plan & 51 & 2.0 & 2.0 & 0.0 \\
Proposed plan validity & 57 & 68.4 & 75.4 & 63.2 \\
\bottomrule
\end{tabular}
\end{table*}

\begin{table*}[!tp]
\caption{\textbf{Annotation-assisted zero-shot results.} Each condition is compared with the same model's replies to the original images on the same eligible test questions. Scores use strict parsing. $n$ gives the eligible question count; a dash indicates an unavailable condition. $^\dagger$The prior-incorrect row is a subset of Component state or screw count.}
\label{tab:suppl_refs}
\centering\footnotesize
\setlength{\tabcolsep}{3pt}
\begin{tabular}{@{}lrrrrrrrrrr@{}}
\toprule
& \multicolumn{5}{c}{State oracle} & \multicolumn{5}{c}{Overlay} \\
\cmidrule(lr){2-6}\cmidrule(l){7-11}
& & \multicolumn{2}{c}{Qwen} & \multicolumn{2}{c}{InternVL} & & \multicolumn{2}{c}{Qwen} & \multicolumn{2}{c}{InternVL} \\
Task & $n$ & Images $\uparrow$ & Oracle $\uparrow$ & Images $\uparrow$ & Oracle $\uparrow$ & $n$ & Images $\uparrow$ & Overlay $\uparrow$ & Images $\uparrow$ & Overlay $\uparrow$ \\
\midrule
\multicolumn{11}{@{}l}{\textbf{Perception}} \\
Visible component boxes & 55 & 2.5 & 0.0 & 0.6 & 0.3 & 55 & 2.5 & 3.0 & 0.6 & 2.4 \\
Component grounding & 47 & 2.1 & 2.1 & 4.3 & 0.0 & -- & -- & -- & -- & -- \\
Component state or screw count & 48 & 39.6 & 83.3 & 43.3 & 83.3 & 54 & 35.2 & 25.9 & 38.5 & 20.4 \\
State or count, prior incorrect$^\dagger$ & 23 & 30.4 & 82.6 & 20.9 & 82.6 & 29 & 24.1 & 20.7 & 16.6 & 10.3 \\
Same step across views, Scanner pairs & 37 & 43.2 & 64.9 & 48.6 & 48.6 & -- & -- & -- & -- & -- \\
Same step across views, OAK-D top pairs & 24 & 58.3 & 100.0 & 58.3 & 83.3 & -- & -- & -- & -- & -- \\
Same step across views, OAK-D side pairs & 12 & 41.7 & 100.0 & 50.0 & 100.0 & -- & -- & -- & -- & -- \\
\addlinespace[2pt]
\multicolumn{11}{@{}l}{\textbf{Understanding}} \\
Observed operation & 54 & 3.7 & 77.8 & 3.7 & 79.6 & 54 & 3.7 & 3.7 & 3.7 & 1.9 \\
Observed operation, region hint & 54 & 3.7 & 74.1 & 3.7 & 77.8 & 54 & 3.7 & 3.7 & 3.7 & 1.9 \\
Operations across several steps & 51 & 10.4 & 61.0 & 5.5 & 65.0 & 51 & 10.4 & 7.7 & 5.5 & 4.6 \\
State change & 74 & 56.8 & 95.9 & 56.8 & 93.2 & 74 & 56.8 & 55.4 & 56.8 & 41.9 \\
State change across several steps & 74 & 60.8 & 94.6 & 56.8 & 94.6 & 74 & 60.8 & 60.8 & 56.8 & 48.6 \\
Completed operations & 58 & 5.2 & 2.7 & 5.2 & 1.7 & 58 & 5.2 & 1.7 & 5.2 & 5.2 \\
\addlinespace[2pt]
\multicolumn{11}{@{}l}{\textbf{Planning}} \\
Proposed action feasibility & 71 & 46.5 & 39.4 & 54.9 & 50.7 & 71 & 46.5 & 32.4 & 54.9 & 52.1 \\
Next feasible action categories & 52 & 3.3 & 11.3 & 2.3 & 9.9 & 52 & 3.3 & 3.8 & 2.3 & 1.2 \\
Next feasible action categories, constrained & 53 & 8.6 & 17.6 & 5.7 & 14.7 & 53 & 8.6 & 9.8 & 5.7 & 4.4 \\
Next feasible action categories, multiple choice & 23 & 73.9 & 34.8 & 0.0 & 26.1 & 23 & 73.9 & 65.2 & 0.0 & 0.0 \\
Target removal plan & 53 & 0.0 & 24.5 & 0.0 & 39.6 & 53 & 0.0 & 0.0 & 0.0 & 0.0 \\
Constrained removal plan & 51 & 0.0 & 3.9 & 0.0 & 2.0 & 51 & 0.0 & 0.0 & 0.0 & 0.0 \\
Proposed plan validity & 57 & 54.4 & 50.9 & 33.3 & 54.4 & 57 & 54.4 & 54.4 & 33.3 & 26.3 \\
\bottomrule
\end{tabular}
\end{table*}

\subsection{Evaluation Sample}
\label{sec:app_vlm_sample}

The vision-language benchmark evaluates a sample of the QA set, drawn once with a fixed seed across all 66 desktops. Questions from validation desktops are used to select checkpoints. The 954 questions from test desktops are used for evaluation. Sampling targets 400 questions for each yes/no task and 300 for each other task. No desktop contributes more than 20\% of a task's target. Two tasks are balanced before sampling. For Component state or screw count, questions are removed until no answer outnumbers all other answers for a desktop and class. For Proposed action feasibility, questions are removed until no single answer accounts for more than 60\% of the questions for any operation--class pair.

Sampling within each task is stratified by factors such as component class, operation, answer, and camera pair. Each multiple-choice question of Next feasible action categories is sampled together with its open form. The multiple-choice form is available for fewer observations, giving 23 multiple-choice questions and 52 open questions on the test desktops. Each sampled Observed operation question includes its region-hint form.

\Cref{tab:level2} reports results for 663 of the 954 test questions, covering the twelve primary tasks. The remaining questions come from the five additional forms and from the OAK-D top and OAK-D side groups of Same step across views (\cref{tab:app_qa_counts}).

\subsection{Models and Fine-Tuning}

Qwen3-VL-8B-Instruct~\cite{bai2025qwen3} and InternVL3.5-8B~\cite{wang2025internvl3} are evaluated zero-shot and after fine-tuning. LoRA~\cite{hu2021lora} is applied to all linear layers of the language model, with rank 16, scaling 32, and dropout 0.05. The vision encoder and projector remain frozen. Training uses the QA pairs from training desktops for two epochs, with AdamW, learning rate $10^{-4}$, cosine decay, 3\% warm-up, weight decay 0.01, and effective batch size 16. The training target is the reference answer in the requested JSON format, without the reasoning trace. Images are resized to at most 1,568 pixels on the longer side.

\begin{figure*}[!tp]
\centering
\includegraphics[width=0.95\textwidth]{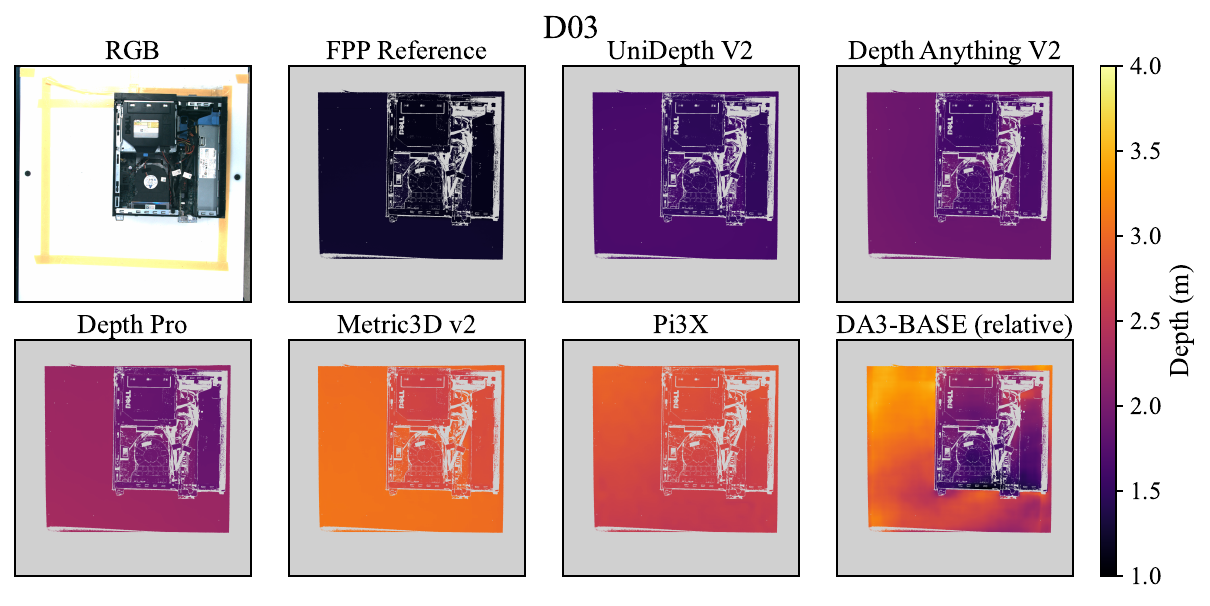}
\caption{\textbf{Qualitative depth comparison (D03, initial observation).} The figure shows the RGB image, FPP reference depth, and predictions from six models. Metric depth maps are displayed over a shared range of 1--4 m without scale alignment. DA3-BASE produces relative depth and is included for visual comparison only.}
\label{fig:depth_comparison}
\end{figure*}

\begin{figure*}[!tp]
\centering
\includegraphics[width=0.9\textwidth]{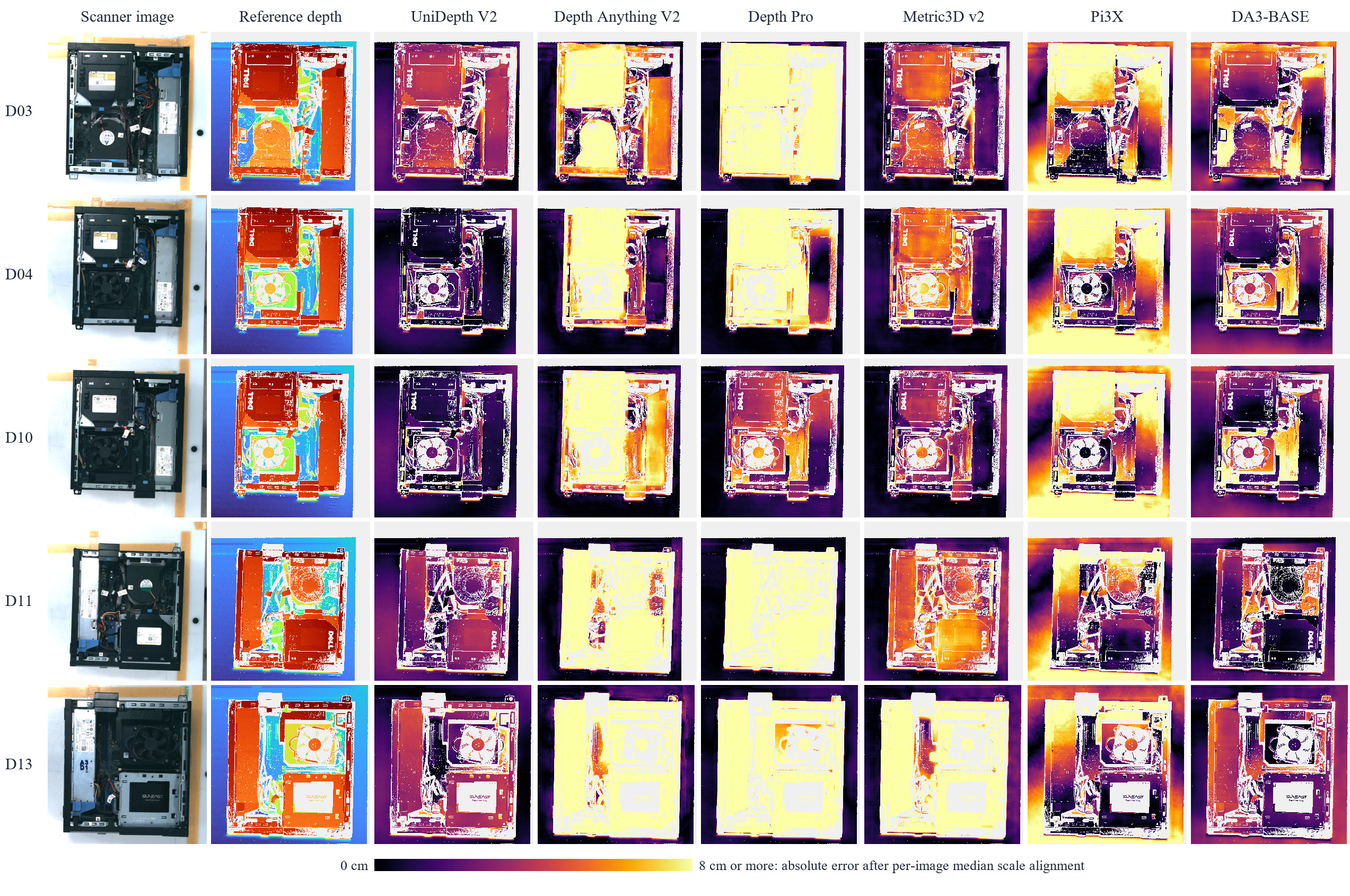}
\caption{\textbf{Depth error on the five evaluation desktops.} Rows show D03, D04, D10, D11, and D13 at their initial observations. Columns show the Scanner image, reference depth projected from the reconstruction, and absolute depth errors for each of the six models. The models are evaluated without fine-tuning. Errors are computed after per-image median scale alignment on pixels with reference depth. Other pixels are light gray. The color scale runs from 0 to 8~cm. Errors of 8~cm or more share the brightest color.}
\label{fig:app_depth_errors}
\end{figure*}

\begin{figure*}[!tp]
\centering
\includegraphics[width=0.8\textwidth]{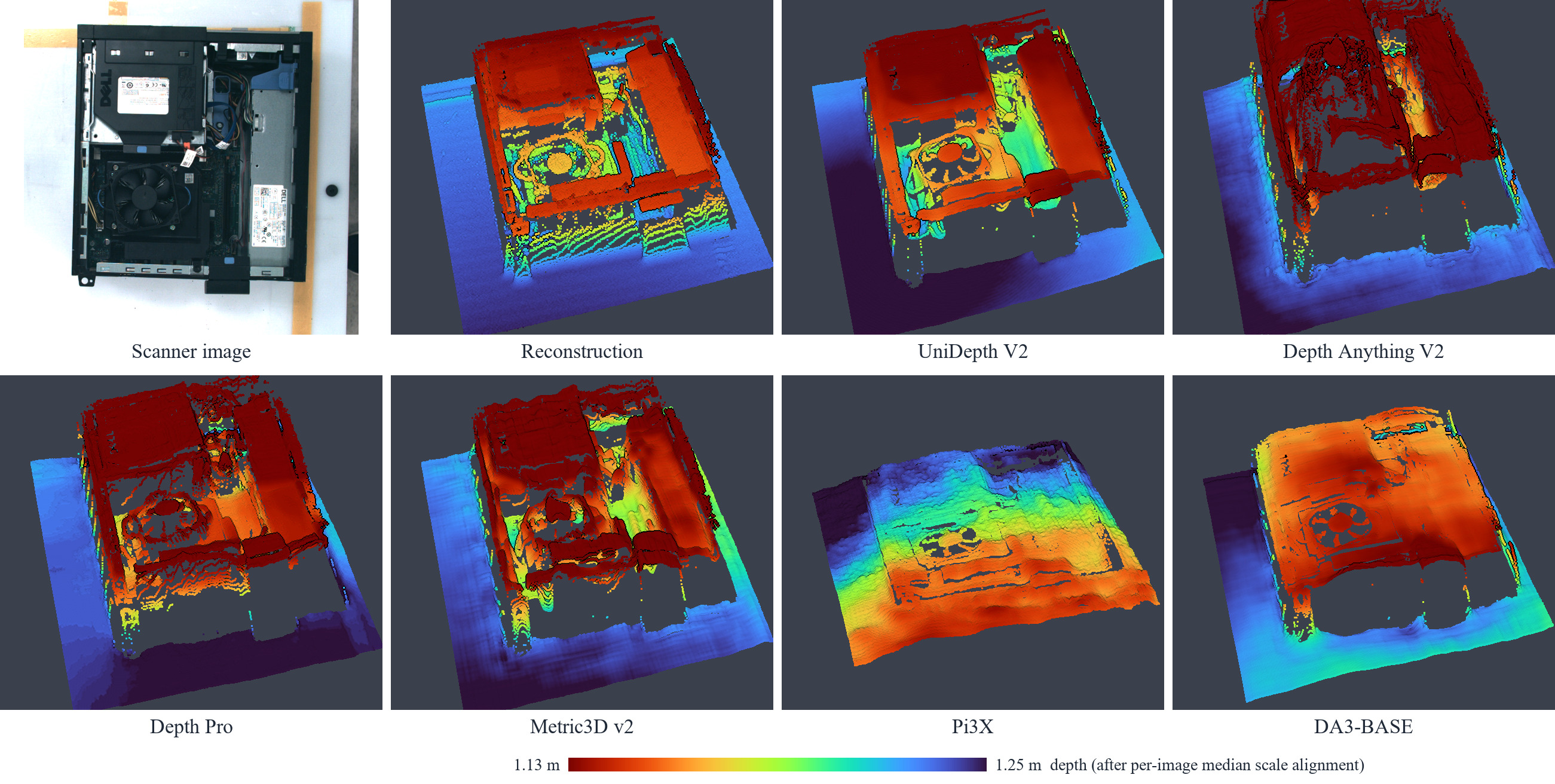}
\caption{\textbf{Predicted depth as point clouds (D10, initial observation).} The models use the initial Scanner image without fine-tuning. Each prediction is aligned by per-image median scaling. The reference depth projected from the reconstruction and each aligned prediction are back-projected with the scanner intrinsics on pixels with reference depth. Points are colored by depth on a shared scale from 1.13 to 1.25~m and rendered from the same oblique viewpoint.}
\label{fig:app_depth_3d}
\end{figure*}

\begin{table*}[!tp]
\caption{\textbf{Robot demonstrations by target component.} A time step has one wrist image and one base image, recorded at 30~Hz.}
\label{tab:app_robot_tasks}
\centering
\small
\setlength{\tabcolsep}{6pt}
\begin{tabular}{@{}lrrrrrr@{}}
\toprule
& \multicolumn{3}{c}{Fixture setting} & \multicolumn{3}{c}{On-motherboard setting} \\
\cmidrule(lr){2-4}\cmidrule(l){5-7}
Target component & Demonstrations & Time steps & Mean length (s) & Demonstrations & Time steps & Mean length (s) \\
\midrule
CPU & 162 & 300,107 & 61.8 & 135 & 242,334 & 59.8 \\
CPU fan & 155 & 114,781 & 24.7 & 131 & 104,572 & 26.6 \\
RAM module & 202 & 139,870 & 23.1 & 160 & 121,263 & 25.3 \\
Connector & 199 & 152,320 & 25.5 & 160 & 137,716 & 28.7 \\
Graphics card & 138 & 155,737 & 37.6 & 130 & 162,276 & 41.6 \\
Hard disk drive & 131 & 140,132 & 35.7 & -- & -- & -- \\
Solid-state drive & 105 & 113,478 & 36.0 & -- & -- & -- \\
\midrule
Total & 1,092 & 1,116,425 & 34.1 & 716 & 768,161 & 35.8 \\
\bottomrule
\end{tabular}
\end{table*}

\begin{figure*}[!tp]
\centering
\includegraphics[width=\textwidth]{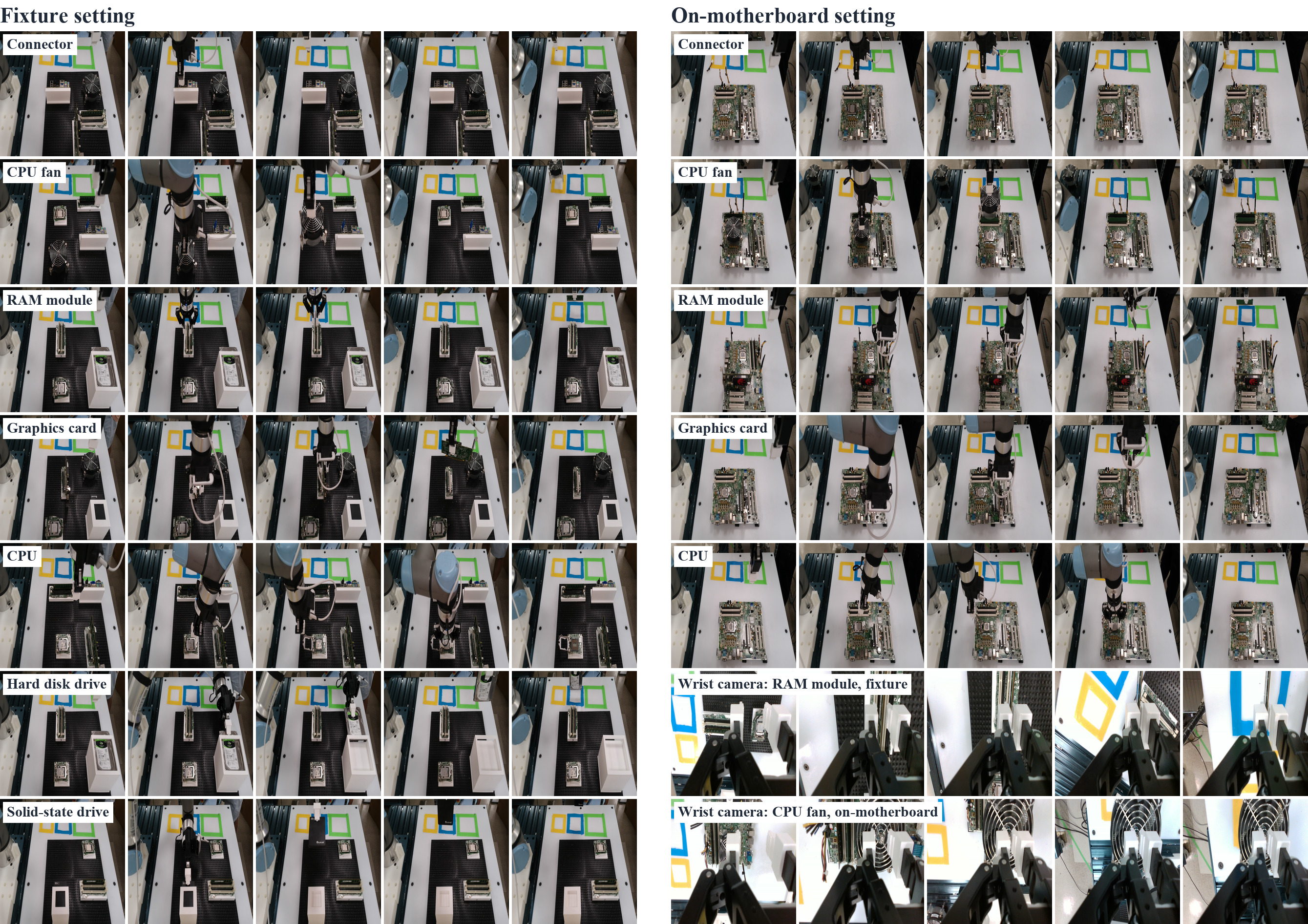}
\caption{\textbf{Robot demonstrations of every target component.} Each row shows frames of one demonstration in temporal order. Left: the seven target components of the fixture setting, from the base camera. Right: the five target components of the on-motherboard setting, from the base camera, and the wrist-camera view of two demonstrations.}
\label{fig:app_robot}
\end{figure*}

After each epoch, the checkpoint is evaluated on the validation questions of the sample. Selection uses the mean score over tasks and forms. Both models select the second epoch. Fine-tuned replies use greedy decoding with at most 2,048 new tokens. Each model has one training run.

GPT-6 Astra and Claude Opus 5.5 use their respective clients at medium reasoning effort. GPT-6 Astra runs through its command-line client in a fresh session per question. Claude Opus 5.5 uses a fresh isolated context per question. Tools are disabled. Temperature and output length use the service defaults. These runs were completed in November 2026. GPT-6 Astra also answers the no-image versions of the planning forms in \cref{tab:suppl_closed954}.

\subsection{Input Conditions and Baselines}
\label{sec:app_vlm_inputs}

The no-image condition removes the images, view legend, and outlined-part list. It also deletes the phrase ``outlined in red'' from questions and options. A target named only by a letter tag keeps the tag, and its class name is added. The system message, answer format, and glossary remain unchanged.

For Target removal plan, 52 of the 53 test questions target the CPU, and 35 share one reference plan. This concentration explains the constant baseline's 66.0 score in \cref{tab:level2}.

\subsection{Results and Parsing Sensitivity}
\label{sec:app_vlm_parsing}

\Cref{tab:suppl_tasks} reports results for all 17 task forms, including comparisons with and without images. Confidence intervals are given in \cref{tab:suppl_intervals}, and closed-model results in \cref{tab:suppl_closed954}. In Next feasible action categories, the multiple-choice form lists candidate operations. The open form supplies neither candidates nor a description of the current assembly. After fine-tuning, Qwen and InternVL obtain the same multiple-choice scores with and without images.

Strict parsing lowers some scores because of the format of a reply rather than its content. It rejects all 23 zero-shot InternVL replies on the multiple-choice form of Next feasible action categories, and lenient parsing recovers 16 correct choices, or 69.6\%. For Proposed plan validity, 20 of 57 InternVL replies fail strict parsing, and accuracy rises from 33.3 to 52.6 with lenient parsing. Across all 954 image replies, 27 Claude Opus 5.5 replies and no GPT-6 Astra replies fail strict parsing. On Component grounding, the Claude Opus 5.5 score rises from 34.9 to 56.2 under lenient parsing, while GPT-6 Astra remains at 52.1.

\subsection{Annotation-Assisted Inputs}

Two further zero-shot conditions provide annotation information as model input (\cref{tab:suppl_refs}). The state-oracle condition replaces the images with tables derived from the annotations of the Scanner images. Each row gives a part's label, state, location (in the chassis or on the workbench), and visibility where available. One table is supplied for each observation in the question, so a change question has a before table and an after table. Outlined targets are marked. No disassembly graph relations, disassembly rules, or operation log entries are added. The question text is transformed as in the no-image condition. All questions are eligible except state questions with the reference answer \texttt{cannot\_tell}, giving 948 of the 954 test questions.

The overlay condition keeps the images and draws the visible-region masks on the Scanner images in a fixed color per class with a thin outline. No legend or text is added. The questions are unchanged. In multi-view inputs, masks are drawn only on the Scanner image. A question is eligible when the Scanner images it shows have annotated masks. Component grounding and Same step across views are excluded.

The state tables improve Observed operation and State change, but neither condition improves Proposed action feasibility for either model.

\section{Monocular Depth Estimation}
\label{sec:app_depth_qual}

The evaluation follows the protocol of the main paper at the $1600\times1600$ resolution of the Scanner image. DA3-BASE predicts at $504\times504$, and its predictions are upsampled to the evaluation resolution. Pi3X is run at an input size close to the evaluation resolution rather than its default of about $504\times504$.

\Cref{fig:depth_comparison} compares the predictions for the initial observation of D03 with the RGB image and reference depth. \Cref{fig:app_depth_errors} shows the absolute depth errors after alignment on all five desktops. \Cref{fig:app_depth_3d} shows point clouds back-projected from the aligned predictions for D10. UniDepth V2 matches the reconstructed component heights most closely, while Depth Anything V2 exaggerates them and DA3-BASE and Pi3X flatten the desktop. Pi3X also tilts the workbench plane.

\section{Robot Demonstrations and Component Disassembly}
\label{sec:app_robot}

\subsection{Demonstration Composition}

The demonstrations are collected with the robot system in the two settings described in the main paper (\cref{fig:overview}). Lighting and background are controlled. \Cref{tab:app_robot_tasks} lists the demonstrations by target component. CPU fans in the table correspond to the CPU cooler annotation class, graphics cards to expansion cards, and hard disk and solid-state drives to storage drives. Each demonstration has a language instruction that names its target. One demonstration of every target component in both settings is shown in \cref{fig:app_robot}.

\subsection{Recorded Signals}

A wrist-mounted RealSense D435i and fixed OAK-D Pro base camera record synchronized $256\times256$ RGB images at 30~Hz (\cref{tab:suppl_hardware}). Each time step also records the joint positions, the tool-center-point (TCP) pose, and the gripper position.

\subsection{Policy Training and Evaluation}
\label{sec:app_robot_evaluation}

OpenVLA-OFT~\cite{kim2025fine} and $\pi_{0.5}$~\cite{intelligence2025pi_} are fine-tuned with LoRA for 50,000 steps on the complete 30~Hz fixture demonstrations of all seven target components. The on-motherboard $\pi_{0.5}$ policy is trained with the same settings on the on-motherboard demonstrations.

In the fixture evaluation, the component positions are chosen at random across the whole base plate. RAM module and connector rollouts are balanced across the available slots. A rollout succeeds when the instructed target is removed and placed in the designated sorting zone. In the on-motherboard evaluation, the positions are also chosen at random across the whole workspace, and the success criterion is the same.

\end{document}